\documentclass[twoside,11pt]{article}

\usepackage{blindtext}
\usepackage{amsmath}

\usepackage[abbrvbib, preprint]{jmlr2e}

\usepackage{xcolor}

\usepackage{graphicx}

\definecolor{vscodegreen}{HTML}{6A9955}

\usepackage{xcolor}
\usepackage{listings}

\lstdefinestyle{mystyle}{
    commentstyle=\color{vscodegreen},
    keywordstyle=\color{blue},
    stringstyle=\color{purple},
    basicstyle=\ttfamily\footnotesize,
    breakatwhitespace=false,         
    breaklines=true,                 
    captionpos=b,                    
    keepspaces=true,                 
    numbersep=5pt,                  
    showspaces=false,                
    showstringspaces=false,
    showtabs=false,                  
    tabsize=2
}

\usepackage{lastpage}
\jmlrheading{23}{2024}{1-\pageref{LastPage}}{1/21; Revised 4/18}{9/22}{21-0000}{Thivin Anandh, Divij Ghose, Himanshu Jain, Sashikumaar Ganesan}

\ShortHeadings{FastVPINNs: Tensor-Driven Acceleration of VPINNs}{Anandh, Ghose, Jain and Ganesan}
\firstpageno{1}

\begin{document}

\title{FastVPINNs: Tensor-Driven Acceleration of VPINNs for Complex Geometries}

\author{\name Thivin Anandh $^1$ \email thivinanandh@iisc.ac.in \\
       \name Divij Ghose $^1$\email divijghose@iisc.ac.in \\
       \name Himanshu Jain $^2$ \email ms19026@iisermohali.ac.in\\
       \name Sashikumaar Ganesan $^{1*}$ \email sashi@iisc.ac.in\\
       \addr $^1$ Department of Computational and Data Sciences\\
       Indian Institute of Science, Bangalore\\
       Karnataka, India \\
       \addr $^2$ Department of Physical Sciences\\
       Indian Institute of Science Education and Research, Mohali\\
       Punjab, India\\
       \addr $^*$Corresponding author}

\editor{My editor}

\maketitle

\begin{abstract}

Variational Physics-Informed Neural Networks (VPINNs) utilize a variational loss function to solve partial differential equations, mirroring Finite Element Analysis techniques. Traditional hp-VPINNs, while effective for high-frequency problems, are computationally intensive and scale poorly with increasing element counts, limiting their use in complex geometries. This work introduces FastVPINNs, a tensor-based advancement that significantly reduces computational overhead and improves scalability. Using optimized tensor operations, FastVPINNs achieve a 100-fold reduction in the median training time per epoch compared to traditional hp-VPINNs. With proper choice of hyperparameters, FastVPINNs surpass conventional PINNs in both speed and accuracy, especially in problems with high-frequency solutions. Demonstrated effectiveness in solving inverse problems on complex domains underscores FastVPINNs' potential for widespread application in scientific and engineering challenges, opening new avenues for practical implementations in scientific machine learning.
\end{abstract}


\section{Introduction}
At present, the realm of applied mathematics is witnessing substantial progress as a result of incorporating deep learning approaches in solving partial differential equations (PDEs). These methods, collectively termed scientific machine learning (SciML) (\cite{cuomo2022scientific, osti_1478744, psaros2023uncertainty}), often complement or replace traditional solvers. Despite the dominance of conventional numerical methods such as the finite element method (FEM), which boasts several open-source solvers (\cite{wilbrandt2017parmoon, bangerth2007deal, ganesan2020sparsh}), SciML has rapidly evolved. This evolution has led to a growing repertoire of accessible SciML libraries (\cite{lu2021deepxde, modulus}).

Since the advent of physics-informed neural networks (PINNs) (\cite{712178, raissi2019physics}), the application of such methods has surged. PINNs enhance the typical data-driven loss function of neural networks by incorporating an additional term that minimizes the residual of the underlying PDE and enforces physics-based constraints. Features such as the ease of obtaining gradients through automatic differentiation, and the ability to train networks for both forward and inverse modeling, make PINNs superior to traditional methods (\cite{abueidda2021meshless, lu2021physics}). Consequently, PINNs have found widespread application in various fields, including solid mechanics (\cite{haghighat2021physics, zhang2022analyses}), fluid mechanics (\cite{mao2020physics, cai2021physics, eivazi2022physics}), and geophysics (\cite{10.1093/gji/ggab309}).

An extension of PINNs, Variational PINNs (VPINNs), utilize the weak form of the PDE in their loss functions, mirroring the Petrov-Galerkin framework used in conventional FEM (\cite{kharazmi2019variational, khodayi2020varnet}). Despite their computational demand due to the necessity to numerically compute integrals in the loss function, VPINNs represent a significant step forward in applying neural networks to PDEs.

Based on VPINN, hp-VPINNs have been developed to improve accuracy through domain decomposition (h-refinement) or increasing polynomial order (p-refinement) (\cite{kharazmi2021hp}). Although hp-VPINNs offer superior accuracy in various problems, they suffer from critical limitations that hinder their practical application. These limitations include their computational complexity, linear scaling of training time with element count, and difficulty in handling complex geometries with skewed elements. These issues are evidenced in both academic and practical scenarios and have been a barrier to greater adoption despite various attempts to deploy hp-VPINNs in different applications (\cite{radin2023effects, yang2021hp}).

Furthermore, recent developments such as convolutional PINNs (cv-PINNs) (\cite{LIU2023102051}) introduced convolutional operations to accelerate the training of hp-VPINNs. Despite claims of superior performance, cv-PINNs still exhibited linear increases in training time with element count in 1D studies, and their solutions have been limited to domains suitable for decomposition into regular quadrilateral cells.

In response to these challenges, this work introduces FastVPINNs, a novel framework designed to address the limitations of hp-VPINNs. The proposed FastVPINNs leverage tensor-based operations to compute the loss, substantially reducing training times and diminishing the impact of increasing element counts. This advancement enables the handling of complex geometries, significantly broadening the practical applicability of our approach.
\section{Preliminaries}
\subsection{Governing Equations}
Consider a two-dimensional steady-state convection-diffusion equation:
\begin{align}
    \begin{split}
        -\varepsilon\Delta u(x) + \mathbf{b}\cdot \nabla u(x) &= f(x), \quad \ \text{in}  \ \ \Omega \subseteq \mathbb{R}^2, \\
        u(x) &=  g(x), \quad \  \text{on} \ \ \partial\Omega.
    \end{split}
    \label{eq:CD2D_strong}
\end{align}
Here, $x \in \Omega $, $\varepsilon$, and $\mathbf{b}$ are the diffusion coefficient and convective velocity, respectively. In addition, $f(x)$ is a known source function with appropriate smoothness. The Dirichlet boundary condition $u(x)=g(x)$ is imposed on the domain boundary $\partial\Omega$. 
The two-dimensional steady-state Poisson equation can be obtained from \eqref{eq:CD2D_strong} by substituting $\mathbf{b}=(0,0)^T$ and $\varepsilon=1$.
\begin{align}
    \begin{split}
        -\Delta u(x) &= f(x), \quad \ \text{in}  \ \ \Omega \subseteq \mathbb{R}^2, \\
        u(x) &=  g(x), \quad \  \text{on} \ \ \partial\Omega.
    \end{split}
    \label{eq:Poisson_strong}
\end{align}
For the remainder of our discussion in this section, we will be using~\eqref{eq:Poisson_strong} as a reference.
\subsection{Physics Informed Neural Networks}
The result generated by a deep neural network is typically structured as a parametric function of the variable $x$, denoted as $u_{\text{NN}}(x; W, b)$. In this context, $W$ and $b$ represent the weights and biases of the network. When the neural network consists of $h$ hidden layers, with the $i^{\text{th}}$ layer containing $n_i$ neurons, the mathematical representation of the function takes the following shape:

\begin{equation}
   u_{\text{NN}}(x; W, b) = l \circ \mathrm{T}^{(h)} \circ \mathrm{T}^{(h-1)} \hdots \circ \mathrm{T}^1(x)
    \label{eq:NN}
\end{equation}
Here, $l : \mathbb{R}^{n_h} \rightarrow \mathbb{R}$ is a linear mapping in the output layer and $\mathrm{T}^{(i)}(\cdot) = \sigma(W^{(i)} \times \cdot + b^{(i)})$ is a non-linear mapping in the $i^{th}$ layer $( i= 1, 2, \cdots , h)$, with the non-linear activation function $\sigma$ and the weights $W^{(i)}$ and biases $b^{(i)}$ of the respective layers.

The neural network's output $u_{\text{NN}}(x; W, b)$ can serve as an approximation to the unknown solution $u(x)$ of the equation~\ref{eq:Poisson_strong} by training the deep neural network. Physics-Informed Neural Networks aim to achieve this estimation by systematically reducing the residual arising from the governing equations specified in equation~\ref{eq:Poisson_strong} with $u_{\text{NN}}(x; W, b)$. Specifically, we refer to the residual related to the partial differential equation (PDE) and the boundary conditions as $\mathcal{P}$ and $\mathcal{B}$, respectively, that is,
\begin{align}
    \begin{split}
        \mathcal{P}(x; W, b) &=  -\Delta u_{\text{NN}}(x; W, b) - f(x), \quad \ \text{in}  \ \  \Omega, \\
        \mathcal{B}(x; W, b) &=u_{\text{NN}}(x; W, b) - g(x), \quad \qquad  \text{on}  \ \  \partial\Omega.
    \end{split} 
\end{align}
The loss function employed in the neural network architecture is designed to integrate two primary components: a Partial Differential Equation (PDE) loss and a boundary loss. These components are defined at particular training points within the domain to reduce inaccuracies associated with the PDE and boundary constraints. This division of the loss function allows for a focused reduction in the difference between the neural network's forecasts and the expected physical phenomena described by the governing equations and boundary constraints. In particular, we define
\begin{align}
    \begin{split}
        L_p(W,b) &= \frac{1}{N_T}\sum_{t=1}^{N_T}\left|\mathcal{P}(x;W,b)\right|^2, \\
        L_b(W,b) &= \frac{1}{N_D}\sum_{d=1}^{N_{D}}\left|\mathcal{B}(x;W,b)\right|^2,\\
        L_{\text{PINN}}(W,b) &= L_p + \tau L_b,
    \end{split}
    \label{eqn:PINNs_Loss_components}
\end{align}
where $N_T$ is the total number of training points in the interior of the domain $\Omega$, $N_D$ is the total number of training points on the boundary $\partial\Omega$ and $\tau$ is a scaling factor applied to control the penalty on the boundary term.

\subsection{hp-Variational Physics Informed Neural Network}
In this section, we initially establish the variational form of the Poisson equation \eqref{eq:Poisson_strong}, followed by introducing the Variational Physics Informed Neural Network. Let $\text{H}^1(\Omega)$ denote the conventional Sobolev space, and define
\[
 V:=\left\{v \in \text{\text{H}}^1(\Omega) : v = 0 \quad \text{on} \quad \partial \Omega \right\}.
\]
The subsequent procedure consists of taking the equation~\eqref{eq:Poisson_strong}, multiplying it by $v\in V$, integrating over $\Omega$, and then utilizing integration by parts on the second derivative term. For more detailed information, please refer to~(\cite{ganesan2017finite}). Consequently, the variational representation of the Poisson equation can be formulated as:\\

\noindent Find $u \in V$ such that,
\[
a(u,v) = f(v) \quad \text {for all } v \in V, 
\]
where
\begin{equation}
    a(u,v) := \int_{\Omega} \nabla u \cdot \nabla v \; dx,  \qquad
        f(v) := \int_{\Omega} fv \; dx.
    \label{eqn:Poisson2D_weakform}
\end{equation}
The domain $\Omega$ is then divided into an array of non-overlapping cells, labeled as $K_k$, where $k=1,2,\ldots,\texttt{N\_{elem}}$, ensuring that the complete union $\bigcup_{k=1}^\texttt{N\_{elem}}K_k = \Omega$ covers the entire domain $\Omega$. In this context, we define $V_h$ as a finite-dimensional subspace of $V$, spanned by the basis functions $\phi_h := \{\phi_j(x)\},\; j = 1,2,\ldots,\texttt{N\_{test}}$, where $\texttt{N\_{test}}$ indicates the total number of basis functions in $V_h$. As a result, the discretized variational formulation related to equation~\eqref{eqn:Poisson2D_weakform} can be written as follows,\\

\noindent Find $u_h \in V_h$ such that,
\begin{equation}
a_h(u_h,v) = f_h(v) \quad \text {for all } v \in V_h,  \label{eqn:Poisson2D_disform}
\end{equation}
where
\[
    a_h(u_h,v) := \sum_{k=1}^{\texttt{N\_{elem}}}\int_{K_k} \nabla u_h \cdot \nabla v \; dK,  \qquad
        f_h(v) := \sum_{k=1}^{\texttt{N\_{elem}}}\int_{K_k} fv \; dK.
\]
These integrals can be approximated by employing a quadrature rule, leading to
\begin{align*}
    \int_{K_k} \nabla u_h \cdot \nabla v \; dK &\approx  \sum_{q=1}^{\texttt{N\_quad}} w_q ~\nabla u_h(x_q) \cdot \nabla v(x_q)\;,  \\
       \int_{K_k} fv \; dK & \approx   \sum_{q=1}^{\texttt{N\_quad}}w_q ~f(x_q)\,v(x_q)\;.
\end{align*}
Here, \texttt{N\_quad} is the number of quadrature points in a cell.

The hp-Variational Physics Informed Neural Networks (hp-VPINNs) framework, as presented by Kharazmi et al. (2021), utilizes specific test functions $v_k$, where $k$ ranges from 1 to N\_elem, that are localized and defined within individual non-overlapping cells across the domain.
\begin{equation}
    v_k= 
    \begin{cases}
      v^p \neq 0, & \text{over $K_k$,} \\
      0, & \text{elsewhere.}
    \end{cases}
    \label{eqn:test_function_summation}
\end{equation}
Here, $v^p$ represents a polynomial function of degree $p$. This selection of test and solution spaces results in a Petrov-Galerkin finite element method. Specifically, $u_h$ is estimated by $u_{\text{NN}}(x;W,b)$, which is the neural network solution, whereas the test function $v_h$ is a predetermined polynomial function. By utilizing these functions, we establish the cell-wise residual of the variational form~\eqref{eqn:Poisson2D_disform} with $u_{\text{NN}}(x;W,b)$ as
\begin{equation}
    \begin{split}
        \mathcal{W}_k(x;W,b)&= \int_{K_k} \left( \nabla u_{\text{NN}}(x;W,b) \cdot \nabla v_k ~ - ~ f\,v_k\right)\,dK.
    \end{split}
\label{eqn:VPINNs_Bound_Loss}
\end{equation}
Further, define the variational loss by
\begin{equation} \label{eq:loss}
    L_v(W,b) = \frac{1}{\texttt{N\_elem}}\sum_{k=1}^{\texttt{N\_elem}}\left|\mathcal{W}_k(x;W,b)\right|^2
\end{equation}
and the cost function of the neural network in  hp-VPINN as 
\begin{equation}
    L_{\text{VPINN}}(W,b) = L_v + \tau L_b.
    \label{eqn:vpinns_final_loss_fn_abbr}
\end{equation}
Here, $L_b$ is the Dirichlet boundary loss as expressed in~\eqref{eqn:PINNs_Loss_components} and $\tau$ is a scaling factor applied to control the penalty on the boundary term.\\

\begin{figure}[ht]
    \centering
    \includegraphics[width=0.99\textwidth]{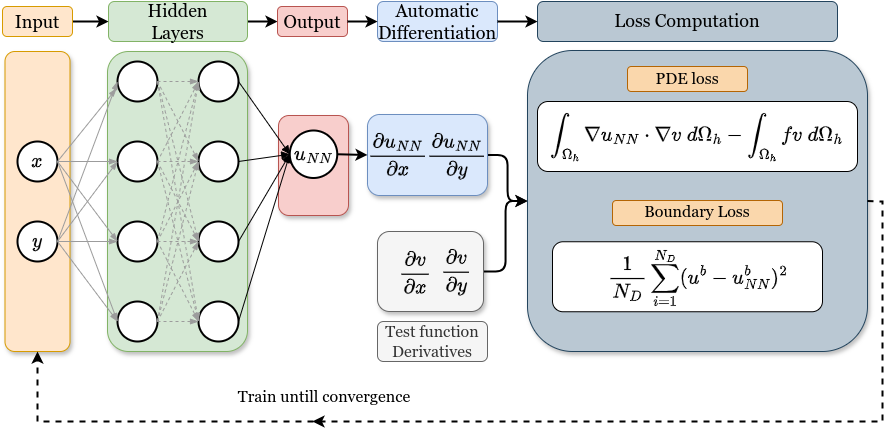}
    \caption{Schematics of Variational PINNs for a 2D Poisson problem.}
    \label{fig:VPINNs schematic}
\end{figure}

\noindent{\textbf{Remark:}} The only distinction between VPINNs and hp-VPINN resides in the selection of the test function. Within the framework of VPINNs, a polynomial that has global support throughout the domain is employed as a test function. Conversely, in the context of hp-VPINN, a compilation of polynomials, each offering support on an element-wise basis, is utilized. Figure~\ref{fig:VPINNs schematic} represents the schematic of VPINNs for a 2D Poisson problem.

\section{Implementation of hp-VPINNs}

This section focuses on the existing implementation of hp-VPINNs. Initially, we explain the current approach to calculating the variational loss in hp-VPINNs using numerical integration methods like Gauss-Lobatto or Gauss-Legendre to achieve accurate integral approximations. Next, we will examine two key limitations of the existing implementation. Firstly, we will explore why the current approach leads to longer training times. Secondly, we will discuss why it may not be suitable for problems involving complex shapes.

\subsection{Overview of Current hp-VPINNs Implementation}
To calculate the cost function related to hp-VPINNs, as specified in Equation~\ref{eqn:vpinns_final_loss_fn_abbr}, it is essential to estimate the integration outlined in Equation~\ref{eqn:VPINNs_Bound_Loss}. This estimation is accomplished by utilizing numerical integration methods, with the Gauss-Lobatto and Gauss-Legendre quadrature methods being particularly effective. As a result of applying these numerical techniques to the integral over the cell $K_k$, $\mathcal{W}_k(x;W,b)$ becomes

\[
\mathcal{W}_k(x;W,b) \approx \sum_{q=1}^{\texttt{N\_quad}}  w_{q}~\left(\frac{\partial u_{\text{NN}}(x_q;W,b)}{\partial x}  \frac{\partial v_k(x_q)}{\partial x}  + \frac{\partial u_{\text{NN}}(x_q;W,b)  }{\partial y}   \frac{\partial v_k(x_q)}{\partial y} - f(x_q)v_k(x_q) \right) \;   
\]
Here, \texttt{N\_quad} denotes the number of quadrature points in a given element $K_k$. 

\noindent\rule{\textwidth}{1pt}  
Algorithm 1: hp-VPINNs Implementation in Kharazmi et.al \newline
\noindent\rule{\textwidth}{1pt}  
\begin{lstlisting}[language=Python]
# VPINNs Existing Implementation
def train_step():
  for k in N_elem:
    u_NN       = model(quadrature_points_in_curr_element) # get NN output
    u_x, u_y   = model.get_grad(u_NN)          # get sol gradients NN
    v_kx, v_ky = fem.get_grad(v_k)             # get test fn grad 
    j,j_x,j_y  = fem.get_jacobians(K_k)        # get jacobians
    w          = fem.get_quad_wt(v_k, K_k)     # get quad wt 
    r[]        = 0
    
    # compute PDE loss
    for q in N_quad: # number of test functions
      for j in N_test: # number of quadrature points
        grad_x = (j[q]/j_x[q])*w[q]*u_x[q]*v_kx[q][j] # du/dx.dv/dx 
        grad_y = (j[q]/j_y[q])*w[q]*u_y[q]*v_ky[q][j] # du/dy.dv/dy 
        r[j]  += (grad_x + grad_y - F[q][j])
        
    # Var loss (F is precomputed outside train loop)
    variational_loss += reduce_mean(square(r))
  
  dirichlet_loss = reduce_mean(square(model(input_bound_pts) - bd_actual)
  total_loss = variational_loss + tau * dirichlet_loss  # Total loss
\end{lstlisting}
\noindent\rule{\textwidth}{1pt}  
\textit{Nomenclature:}~\\ \texttt{v\_kx, v\_ky}: gradients of the test functions in the x and y directions at the k-th cell.  \texttt{u\_x, u\_y}: gradients of \texttt{u\_NN} in the x and y directions (computed using AutoDiff).  \texttt{j[q]}: Jacobian of the element at the specified quadrature point. \texttt{j\_x[q]}, \texttt{j\_y[q]}: Jacobian of the element at the specified quadrature point in the x and y directions.

\noindent\rule{\textwidth}{1pt}  

Algorithm~1 describes the approach for quantifying the loss function in the context of hp-VPINNs, following the guidelines provided in the GitHub repository (\cite{hp_vpinns_github}). This approach involves an iterative procedure that covers each element of the computational domain to compute the local PDE loss. Local losses are then combined to obtain the overall PDE loss measure. Central to this method is the need for gradients and function values of the test functions, which are obtained from predefined basis functions such as Jacobi polynomials. Simultaneously, the neural network provides solutions and gradients at specified points within each element. After computing the local losses, the algorithm consolidates these losses from all elements and combines them with the prescribed boundary loss, as detailed in equation~\eqref{eqn:VPINNs_Bound_Loss}. This unified loss function is crucial for training the neural network, ensuring that the hp-VPINNs model is finely adjusted not only to the underlying physics described by the PDE but also to the specified boundary constraints.
 
\subsection{Need for Efficient hp-VPINNs Implementation}

hp-VPINNs employs h-refinement that restricts the test functions to smaller areas within the computational domain. This restriction of test functions allows hp-VPINNs to capture high frequency features in the solution. However, increasing the number of elements leads to a proportional increase in the time required to train the model.

\begin{itemize}
    \item \textbf{Training time complexity}: The training time of hp-VPINNs increases linearly with the addition of more elements, even when the total number of quadrature points remains constant throughout the domain, as demonstrated in Figure~\ref{fig:hpvpinns_time}. Consequently, employing h-refinement, which involves adding more elements to handle rapidly changing solutions, does not provide the anticipated advantage, as it substantially increases the computational cost.
    \begin{figure}[h!]
    \centering
    \includegraphics[width=0.85\linewidth]{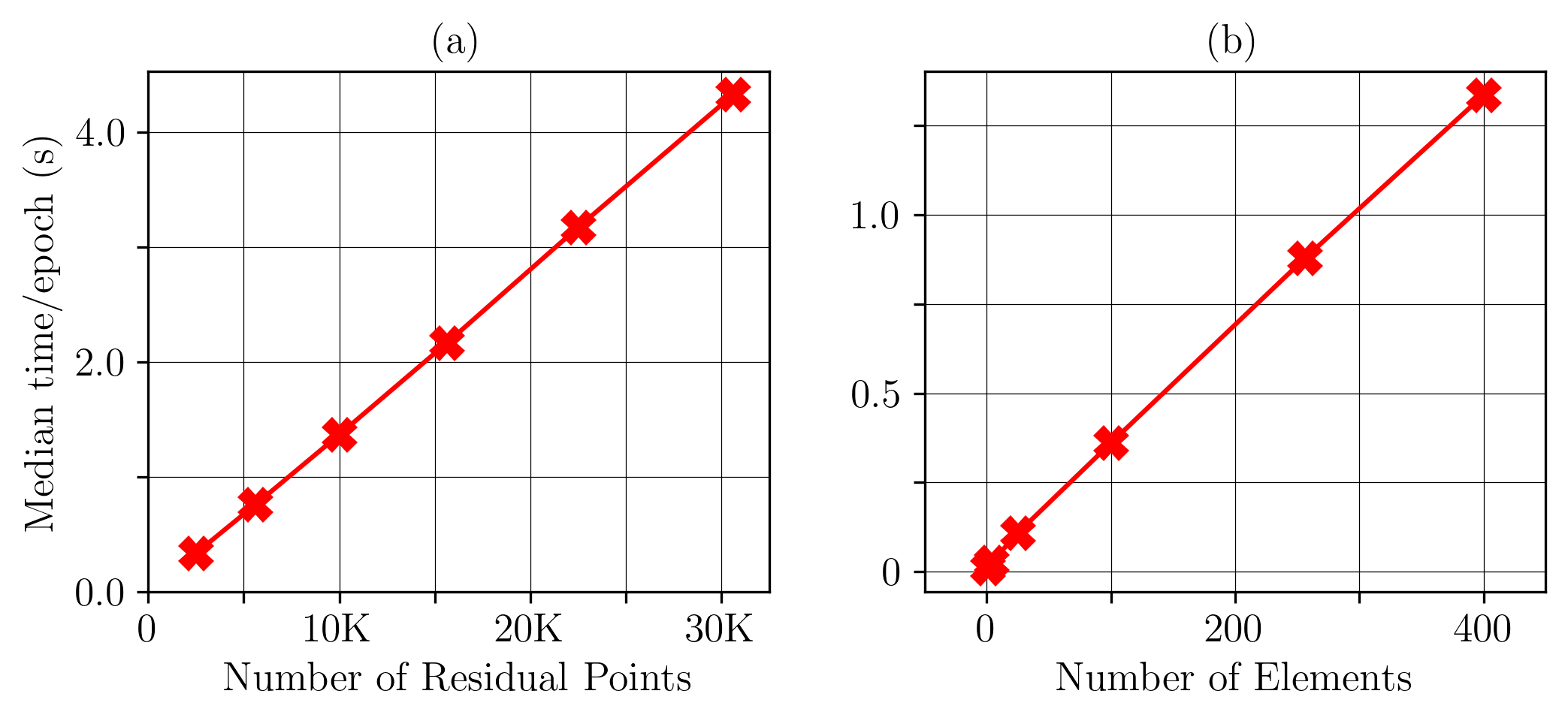}
    \caption{\textbf{hp-VPINNs: Effect of training time on Number of elements} (a) Number of residual points vs training time.(Each element has $25$ quadrature points, so 30K quadrature points will have 1200 elements).(b) Number of elements vs training time(total number of quadrature points within the domain is kept constant at $6400$).}
    \label{fig:hpvpinns_time}
\end{figure}
\item\textbf{Handling complex geometries}: The current version of hp-VPINNs is limited to operate with structured quadrilateral elements. This limitation stems from the current implementation using constant Jacobian values for each element to transform the gradients from the reference element to the actual element. However, practical applications often require skewed elements within complex domains as shown in Figure~\ref{fig:gear_mesh}), where the Jacobian will not be constant within an element.

    \begin{figure}[h]
    \centering
    \includegraphics[width=0.8\textwidth]{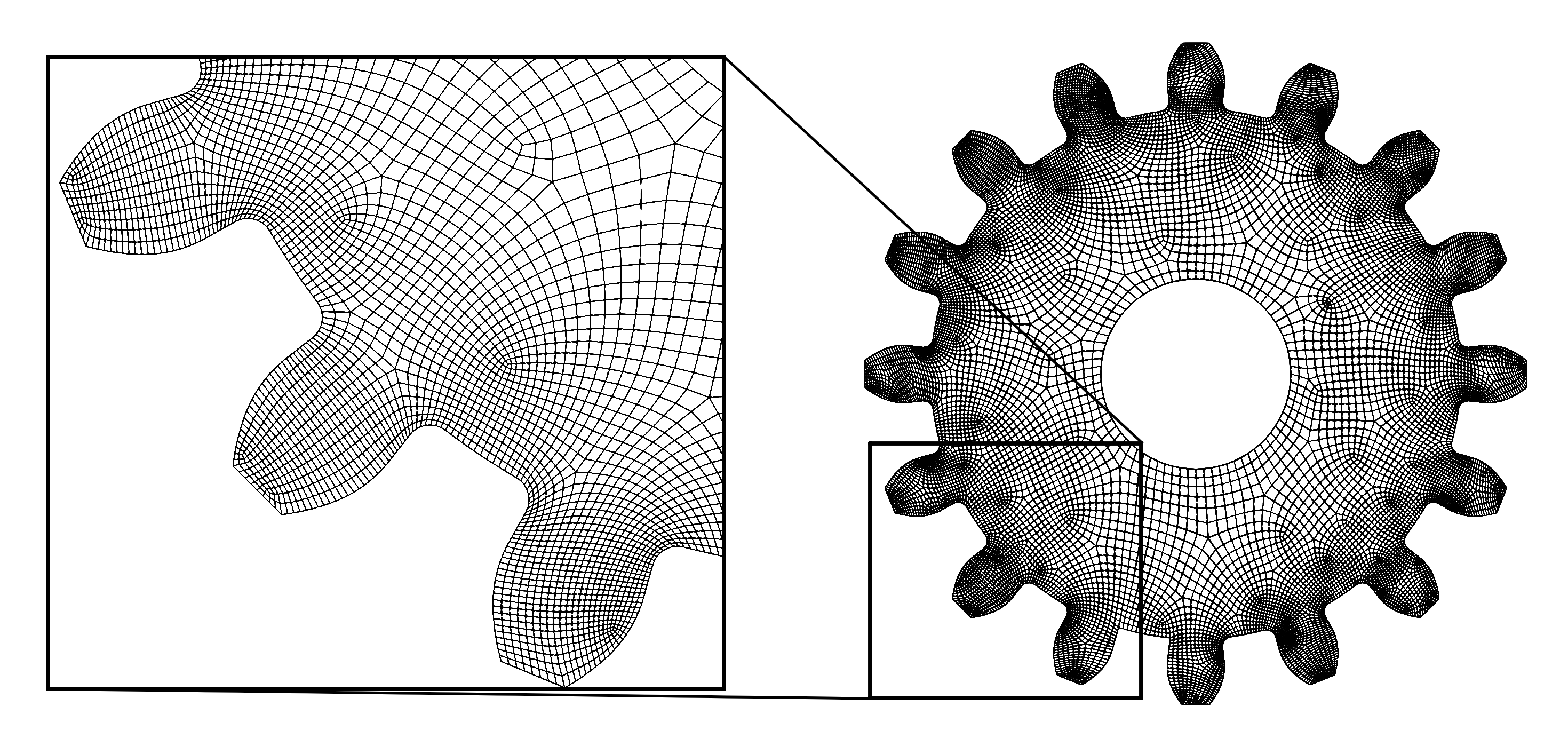}
    \caption{Mesh of spur gear with 14,000 quad elements.}
    \label{fig:gear_mesh}
\end{figure}
\end{itemize}

\section{FastVPINNs Methodology}

This section introduces FastVPINNs, a novel framework that significantly improves upon the existing hp-VPINNs. FastVPINNs tackles two critical challenges: handling complex geometries and achieving faster training times. We achieve these advancements through several key optimizations, including eliminating element-wise looping, removing redundant computations, and leveraging BLAS (Basic Linear Algebra Subprograms) routines. For handling complex geometries, we incorporate concepts of mapped finite elements by utilizing bilinear transformations.

\subsection{Mapped hp-VPINNs}

\begin{figure}[ht]
    \centering
    \includegraphics[width=0.5\textwidth]{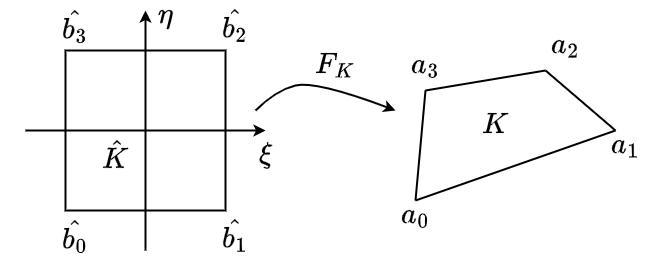}
    \caption{Bilinear transformation.}
    \label{fig:bilinear Transformation}
\end{figure}

Complex geometric shapes frequently lead to the presence of irregular triangles or quadrilaterals in 2D, which pose challenges when it comes to accurately computing integrals and derivatives. To tackle this challenge, the FEM employs mapped finite elements. This approach converts all irregular elements in the actual domain into simpler shapes within a common reference element, as shown in Figure~\ref{fig:bilinear Transformation}. 

The adoption of mapped finite elements offers several advantages:
\begin{itemize}
    \item The simplification of integrals and derivatives is achieved by moving these calculations to a reference element.
    \item The need to compute gradients multiple times for each element is eliminated, as the reference gradients are consistent across all elements in the domain.
\end{itemize}
In this work, the bilinear transformation is employed when dealing with quadrilateral elements. Further details on bilinear transformation can be found in the Appendix.

\subsection{Optimization I: Enhancing Efficiency with Matrix-Vector Product Reformulation}
Upon a detailed analysis of Algorithm~1, it is evident that its computational procedure consists mainly of iterative multiplication operations. These operations pertain to the derivatives of test functions and the quadrature weights utilized for numerical integration, which remain constant throughout the training cycle. The only component that undergoes variation at each training step is the predicted solution and its gradients derived from the neural network.  Recognizing this recurrent pattern, we've devised an approach to modify the gradient computation, denoted as \texttt{grad\_x} in Algorithm 1, to be expressed as a matrix-vector product. Initially, a matrix is prepared in advance, known as the precomputed test function matrix. This matrix, sized $\texttt{N\_test} \times \texttt{N\_quad}$, comprises pre-computed values obtained by multiplying each shape function by the corresponding quadrature weights. During the training process, this matrix is multiplied by the solution of the neural network and a scaling factor known as Jacobian. This multiplication yields a vector of residuals, with dimensions $\texttt{N\_test} \times 1$. This has the following advantages.

\begin{itemize}
    \item \textbf{Using BLAS for Faster Calculations:} By organizing our loss computations in a matrix-vector format, we can utilize the BLAS (Basic Linear Algebra Subprograms) routines. These optimized routines are designed to operate efficiently on GPUs, enhancing the speed of our calculations.

    \item \textbf{Elimination of Redundant Calculations:} Since the test function matrix is precomputed and remains unchanged during training, except for cases such as moving domain problems where the domain shape changes, it is not necessary to repeatedly compute the product of the derivatives of the shape functions and the quadrature weights at each training iteration. This efficient approach saves considerable time and computational resources, thereby improving the speed and effectiveness of the training process.
\end{itemize}

\subsection{Optimization II: Overcoming the Element Looping for regular elements}
\begin{figure}[h]
    \centering
    \includegraphics[width=0.8\textwidth, height=0.5\textheight, keepaspectratio]{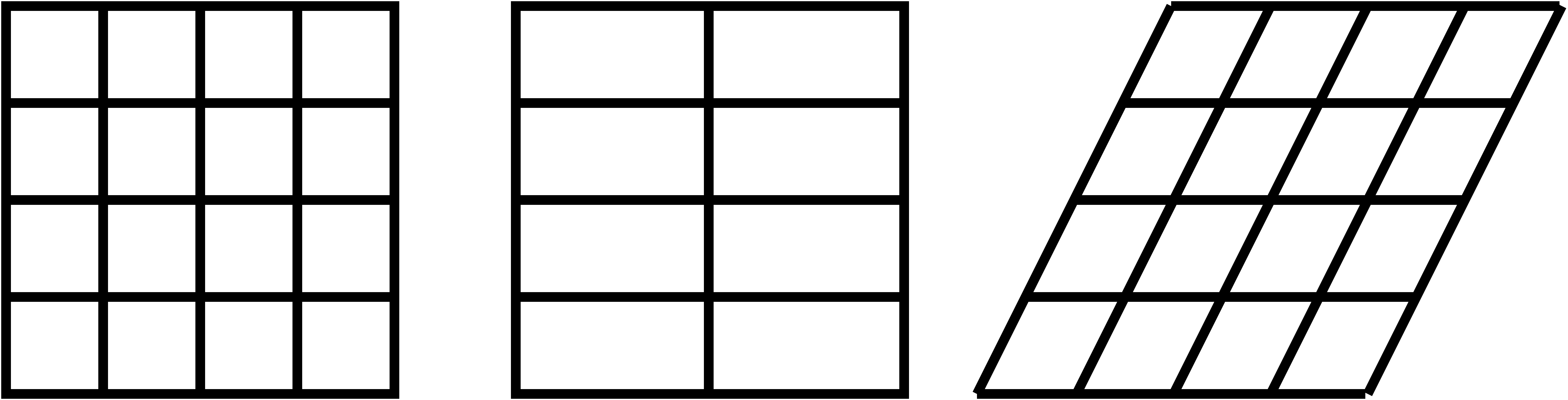}
    \caption{Quadrilateral meshes with elements having constant Jacobian.}
    \label{fig:meshes}
\end{figure}

Despite the efforts made to reduce redundant computations and optimize matrix vector operations, our algorithm still requires the individual processing of each element for the calculation of loss within the computational domain. This means that during every training iteration, the neural network must calculate the solution \texttt{N\_elem} times (known as forward passes) and then compute the gradients of these solutions another \texttt{N\_elem} times (known as backward passes) to determine the loss. This sequential process is a major factor that contributes to the slowdown in the training of hp-VPINNs. For basic non-distorted quadrilateral elements (similar to those shown in Figure~\ref{fig:meshes}), the Jacobian values are constant for a given element. Instead of computing the gradients individually for each element, a single reference gradient matrix can be computed and later multiplied with the corresponding Jacobians to obtain the actual gradients for each element.

In our approach, we exploit these characteristics of a domain composed of regular elements to perform loss computation without the need to iterate through individual elements. The process involves gathering quadrature points, which are used for numerical integration, from all elements across the domain. These points are organized into a single input of dimensions (\texttt{N\_quad} x \texttt{N\_elem}, 2) and then supplied as input to the neural network.Consequently, the neural network produces an output in the form of a vector that can be reorganized into a matrix of size \texttt{N\_quad} $\times$ \texttt{N\_elem}. Each column of this matrix represents the predicted results at the quadrature points for each element.  Subsequently, we proceed by taking each column of this matrix and multiplying it by the corresponding Jacobian. The loss calculation now involves multiplying the premultiplier matrix (reference gradient) with the gradient matrix from the neural network, which is scaled by the Jacobian. The result of this multiplication produces a final matrix with dimensions $\texttt{N\_test} \times \texttt{N\_elem}$, where each column represents the residuals computed for each element within the actual domain.

The key advantages of this optimized approach include:
\begin{itemize}
    \item \textbf{Single Backpropagation per Iteration:}  Through the aggregation of inputs from all elements, we are able to compute the gradients for all elements simultaneously, requiring only a single forward pass and one backward pass (for gradient calculation) across the neural network. 
    
    \item \textbf{Elimination of Element Looping:}  By combining all inputs and using matrix multiplication, we avoid individually processing each element. This approach significantly accelerates our computations and enhances the overall computational efficiency.
\end{itemize}

\noindent\rule{\textwidth}{1pt}  
Algorithm 2: Vectorised hp-VPINNs  with elimination of element looping \newline
\noindent\rule{\textwidth}{1pt}  
\begin{lstlisting}[language=Python]
# J_x, J_y: premultiplier jacobian matrices (N_quad, N_elem)
# F       : preassembled force function matrix (N_elem, N_test)
# V       : test function matrix (N_test, N_quad)
# V_x, V_y: test function gradient matrices (N_test, N_quad)
# NOTE    : refer Appendix for pseudocode of pre-assembly matrices.
def train_step():
  u_NN = model(quadrature_points_in_curr_element)    # get NN output
  
  # get gradients and reshape them
  u_x, u_y = model.get_gradients().reshape(N_quad, N_elem)
  
  # Multiply jacobians of corresponding elements to sol gradients
  # Perform mat-mat multiplication 
  grad_x = matmul(grad_x_mat, u_x * J_x) # int(du/dx.dv/dx),
  grad_y = matmul(grad_y_mat, u_y * J_y) # int(du/dy.dv/dy)
  
  # obtain mse for each element by row reduction (axis 0)
  residual_elements = reduce_mean(square(grad_x + grad_y - F), axis=0)
  variational_loss = reduce_sum(residual_elements) # sum all residuals
  total_loss = variational_loss + beta * dirichlet_loss   # Final loss
\end{lstlisting}
\noindent\rule{\textwidth}{1pt}  

Algorithm 2 details a method for stacking input tensors to compute all gradients in a single backpropagation step, enabling efficient loss calculation for a domain composed of regular elements. The assembly process for the premultiplier matrices is described in the Appendix.

\subsection{FastVPINNs: Generalized algorithm for Complex Geometries}
Though the method we mentioned earlier results in notable speed enhancements, it operates under the premise that the Jacobian remains constant for every element. While this assumption is true for regular elements, it becomes less applicable when dealing with skewed quadrilaterals. In such scenarios, the Jacobian may vary at distinct integration points (quadrature points) and for each shape function.

To effectively tackle this challenge, it is essential to have a unique pre-multiplier matrix which stores the the gradients of the actual elements after transformation. However, using this strategy leads to a process that requires looping through each element to calculate loss, similar to the methodology outlined in Algorithm 1. This renders the previous optimization unfeasible for domains with skewed elements.

\begin{figure}[h]
    \centering
    \includegraphics[width=0.8\textwidth]{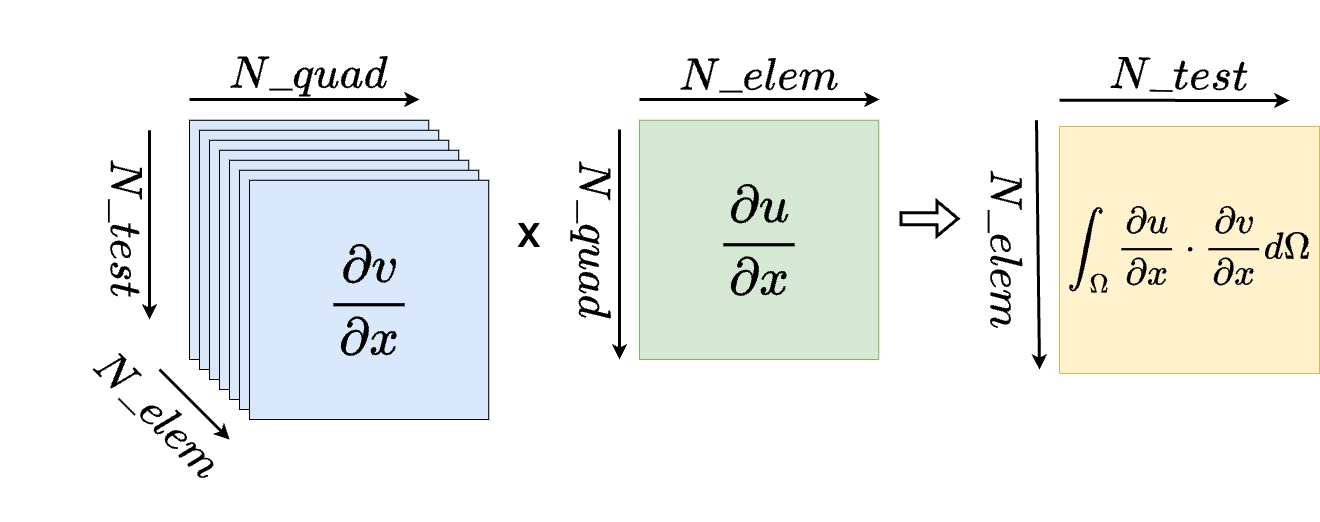}
    \caption{FastVPINNs Tensor schematic representation for residual computation.}
    \label{fig:fast_vpinns}
\end{figure}

In this section, we present a novel approach known as FastVPINNs. This technique stacks the pre-multiplier matrices into a three-dimensional array, or third-order tensor, with dimensions $\texttt{N\_elem} \times \texttt{N\_test} \times \texttt{N\_quad}$. The neural network solutions are structured into a two-dimensional matrix sized $\texttt{N\_quad} \times \texttt{N\_elem}$. This configuration facilitates specific operations using tensors, where each layer of the tensor (representing each element) is systematically multiplied by the corresponding column of the matrix of solution gradients. The result is a set of residual vectors, each of size $\texttt{N\_test} \times 1$ for every element. These vectors populate the columns of a final residual matrix sized $\texttt{N\_test} \times \texttt{N\_elem}$, as illustrated in Figure~\ref{fig:fast_vpinns}.
However, to calculate residuals element-wise using basic linear algebra subprograms (BLAS) such as tensor-matrix multiplication, it is necessary to transpose the dimensions of the solution gradients to $\texttt{N\_elem} \times \texttt{N\_quad}$. This adjustment enables the utilization of TensorFlow's~\cite{tensorflow2015-whitepaper} \texttt{tf.linalg.matvec} function for efficient multiplication of tensor-matrix, resulting in a residual matrix of size $\texttt{N\_elem} \times \texttt{N\_test}$. Subsequently, this matrix can be transposed to achieve the desired layout of $\texttt{N\_test} \times \texttt{N\_elem}$.

The FastVPINNs approach offers several significant benefits:

\begin{itemize}
    \item \textbf{Handles Complex Geometries:} Through the manipulation of tensors, this approach eliminates the necessity of iterating through individual elements, even when working with complex geometries.
    \item \textbf{Efficient Loss Computation with BLAS:} The loss computation, being formulated as tensor-based operations, is well-suited for GPU computations using BLAS routines. Additionally, GPUs typically have tensor cores designed specifically for tensor-based calculations, allowing these routines to leverage optimized hardware capabilities.
    
    \item \textbf{Requires Only One Backpropagation Pass:} In contrast to existing approaches that require multiple backpropagation to compute gradients, FastVPINNs accomplishes this in a single step even for complex geometries.
    
\end{itemize}

\newpage
\noindent\rule{\textwidth}{1pt}  
Algorithm 3: Generalised Tensor-based FastVPINNs Algorithm \newline
\noindent\rule{\textwidth}{1pt}  
\begin{lstlisting}[language=Python, label={lst:algorithm_3}]
# Compute the prematrix multipliers for testFunction and gradients 
# Obtain test fn and gradients in actual element and stack them 
# to form a tensor of shape N_elem x N_test x N_quad
test_tensor   =  tf.stack(test_fn, axis=0)
grad_x_tensor =  tf.stack(grad_x_mat_list, axis=0)
grad_y_tensor =  tf.stack(grad_y_mat_list, axis=0)

# Compute Forcing Matrix similar to Algorithm 2

def train_step():
    # obtain solution for the quadrature points in entire domain
    sol = model(quad_points_in_domain)
    
    # obtain gradients and reshape them
    u_x, u_y = model.get_gradients().reshape(N_elem, N_quad)

    # Perform tensor-matrix mult to evaluate the integral
    grad_x = tf.transpose(tf.matvec(grad_x_tensor, u_x))
    grad_y = tf.transpose(tf.matvec(grad_y_tensor, u_y))

    # Subract the forcing matrix from diff to get residual 
    residual_matrix = grad_x + grad_y - F 

    # obtain mse error for each element in domain
    residual_elements =  reduce_mean(square(residual_matrix), axis=0)
    # sum residuals of all elem in domain
    variational_loss = reduce_sum(residual_elements)
    total_loss = variational_loss + beta * dirichlet_loss

\end{lstlisting}
\noindent\rule{\textwidth}{1pt} 

Algorithm~3 summarizes the approach for FastVPINNs framework. In the next section, we assess the performance of the proposed FastVPINNs framework.
\subsection{Experimental Setup and Design}
In the present research, every experiment was conducted on a system that boasts an NVIDIA RTX A6000 GPU with 48 GB of device memory. The NVIDIA-Modulus library~\citep{nvidia_modulus_docs} was used for physics-informed neural networks (PINN). To establish a baseline comparison with hp-VPINNs, we use the cutting-edge hp-VPINNs code available on GitHub~\citep{hp_vpinns_github}. To ensure uniformity in all our experiments, we opted for the same kind of Jacobi polynomials as outlined in~\citep{kharazmi2021hp}. These polynomials, denoted as $P_{n}$ for degree $n$, follow the equation $P_{n} = P_{n+1} - P_{n-1}$. We also integrated the Gauss-Jacobi-Lobatto quadrature for numerical integration in our computations. The hp-VPINNs software generally functions using 	exttt{tf.float64} precision for floating-point computations. However, our code is compatible with both 	\texttt{tf.float64} and the slightly less accurate 	\texttt{tf.float32}. In all our experiments, we employed 	\texttt{tf.float32}, except when stated otherwise. The duration of each experiment was recorded using the 	\texttt{time.time()} function in Python. In the following sections, the phrase ``residual points'' will refer to the collocation points in PINNs and the quadrature points used in hp-VPINNs and FastVPINNs..

Initially, we conduct a comparison between the accuracy of FastVPINNs and PINNs, and delve into the impact of h- and p-refinement on the convergence of FastVPINNs in Section~\ref{sec:acc_fast_vpinn}. The duration of training of PINNs, hp-VPINNs, and FastVPINNs is compared in Section~\ref{sec:speed_fast_vpinn}. The efficacy of PINNs and FastVPINNs in dealing with high-frequency solution problems is examined in Section~\ref{sec:spectral_fast_vpinn}. The capability of FastVPINNs to tackle forward problems in intricate geometries is exhibited in Section~\ref{sec:complex_fast_vpinn}. Lastly, the performance of FastVPINNs in resolving inverse problems in intricate geometries is probed in sections~\ref{sec:inverse_fast_vpinns} and~\ref{sec:domain_inverse_fast_vpinns}.

\subsection{Evaluation of FastVPINNs}
\begin{figure}[h!]
    \centering
    \includegraphics[width=\linewidth]{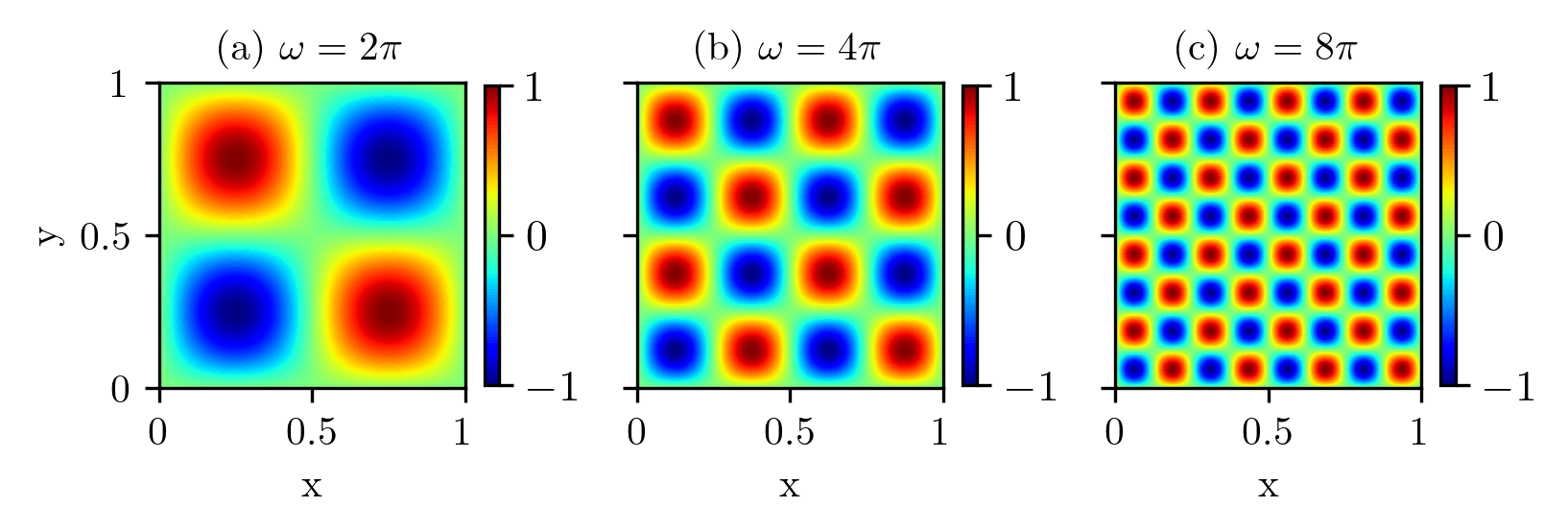}
    \caption{Exact solutions of the two-dimensional Poisson's equation for given test-cases.}
    \label{fig:exact_solution_poisson}
\end{figure}

Initially, we evaluate our code's efficacy on forward problems by solving the two-dimensional Poisson's equation within the unit square, using the specified forcing function

\begin{align*}
    \begin{split}
        -\Delta u(x, y) &= -2\omega^2\sin{(\omega x)}\sin{(\omega y)} \quad \text{in} \in \Omega = (0, 1)^2,\\
        u(0,\cdot) &= u(\cdot,0) = 0.
    \end{split}
\end{align*}
This problem has the exact solution,
\[
    u(x,y) = -\sin(\omega x)\sin(\omega y).
\]
To evaluate the effectiveness of the suggested solver in different solution frequencies, we run the code with $\omega = 2\pi, 4\pi$ and $8\pi$. The exact solutions for these cases are illustrated in Figure~\ref{fig:exact_solution_poisson}.

\subsubsection{Accuracy Test}\label{sec:acc_fast_vpinn}
In this section, we evaluate the accuracy of the proposed FastVPINNs framework by comparing it with the conventional PINNs technique on a problem where the parameter $\omega$ is set to $2\pi$. This comparative analysis involves both methodologies, using neural networks composed of three hidden layers, each containing 30 neurons. The FastVPINNs solution is derived using a grid of $2 \times 2$ elements, each element encompassing $40 \times 40$ quadrature points and 15 test functions per direction. The PINNs framework is trained using a total of $6400$ collocation points, which corresponds to the total number of quadrature points employed in FastVPINNs. Both methodologies were trained for 100,000 iterations. The accuracy of their results was evaluated using a grid of $100 \times 100$ uniformly distributed points.

\begin{figure}[h]
  \centering
  \includegraphics{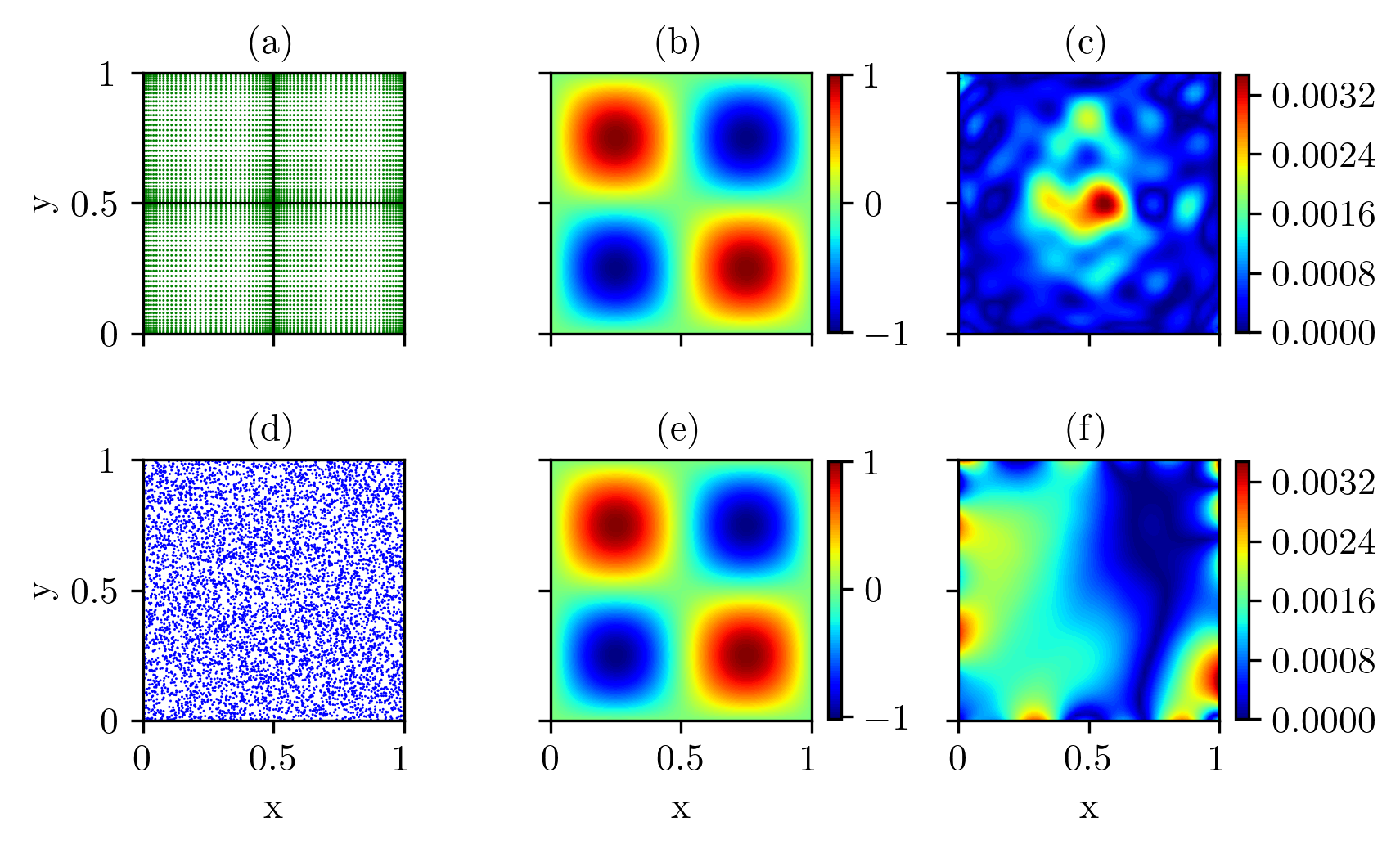} 
  \caption{Accuracy comparison between PINNs and FastVPINNs. The quadrature points used to train FastVPINNs are shown in (a). The FastVPINNs prediction is shown in (b) and the point-wise error for the FastVPINNs solution is shown in (c). The PINNs is trained on collocation points as shown in (d). The PINNs solution and point-wise error are shown in (e) and (f), respectively.}
  \label{fig:vpinn-intro}
\end{figure}

As illustrated in Figure~\ref{fig:vpinn-intro}, the results show that FastVPINNs can reach a comparable level of accuracy as traditional PINNs. Furthermore, it is highlighted that the accuracy of FastVPINNs can be enhanced via h- and p-refinement.

\begin{figure}[h!]
    \centering
   \includegraphics[width=0.8\linewidth]{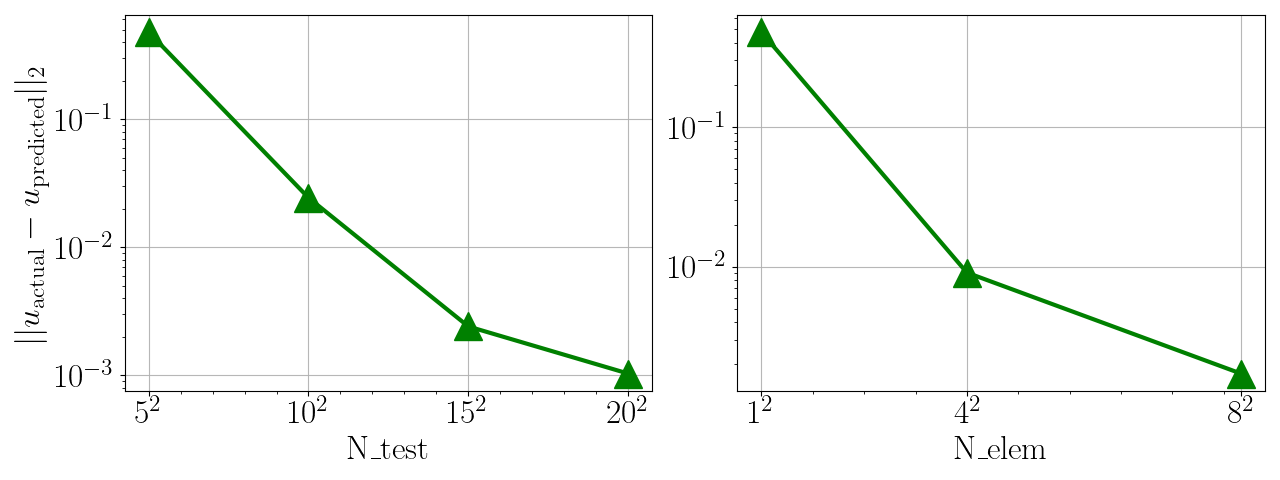}
    \caption{Left: Effect of p-refinement(left) and h-refinement(right) on FastVPINNs.}
    \label{fig:h-p-refinement}
\end{figure}

The impact of h-refinement is examined using an example where $\omega=4\pi$. The process begins with a single element, after which the domain is partitioned into $4\times 4$ grids, and then into $8\times 8$ grids, while keeping the number of quadrature points per element constant at $80\times 80$ and using five test functions in each direction within each element. The model has difficulty accurately depicting the shape or magnitude of the solution when only one element is used. However, when the grid is expanded to $8 \times 8$ elements, the error is significantly reduced to approximately $\mathcal{O}(10^{-3})$, as illustrated in Figure~\ref{fig:h-p-refinement}(a). In the case of p-refinement, increasing the number of test functions in a $1\times 1$ element grid from $5\times 5$ to $20\times 20$ reduces the error from $\mathcal{O}(10^0)$ to $\mathcal{O}(10^{-2})$, as depicted in Figure~\ref{fig:h-p-refinement}(b), thereby proving the effectiveness of p-refinement. A detailed error analysis for both h-refinement and p-refinement is provided in the Appendix~\ref{sec:appendix_hp_refinement}.

\subsubsection{Efficiency Test: Comparison of FastVPINNs with PINNs and Hp-VPINNs}\label{sec:speed_fast_vpinn}
\begin{figure}[h!]
  \centering
  \includegraphics[width=\textwidth]{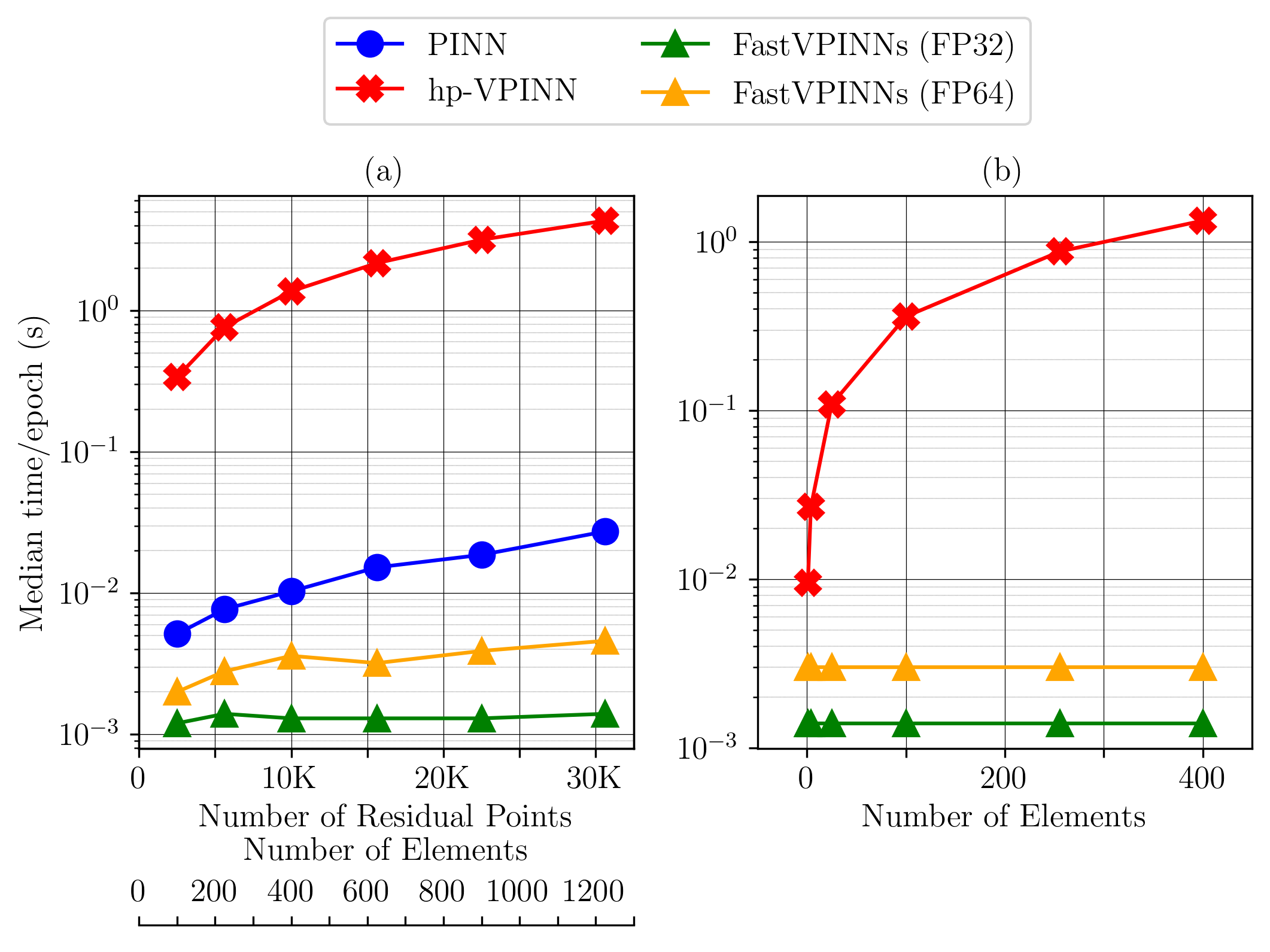} 
  \caption{(a) Variation of computational time with the number of quadrature (residual) points, plotted against the median time taken per epoch; (b) Comparison of computational time between hp-VPINNs and FastVPINNs for varying numbers of elements.
}
  \label{fig:pinn_vs_fastvpinn_time}
\end{figure}

In this section, we assess the performance of our FastVPINNs framework by comparing it with two alternative methods: PINNs and the original version of hp-VPINNs. To measure computational proficiency, we recorded the duration needed to finish one training cycle for each framework. The evaluations were performed over 1,000 cycles to determine the median duration consumed. The results of the comparision are portrayed in Figure~\ref{fig:pinn_vs_fastvpinn_time}(a), where the x-axis indicates the count of residual points used for the calculations (collocation points for PINN and quadrature points for hp-VPINN) in addition to the total count of elements taken into account. For both hp-VPINNs and FastVPINNs, we applied five test functions on the x- and y-axes, culminating in a sum of 25 quadrature points per element and 25 test functions in total. This experimental arrangement facilitates a straightforward efficiency comparison between FastVPINNs and PINNs, using the count of collocation points (quadrature) as a shared factor. The results depicted validate that FastVPINNs surpass both PINNs and the original version of hp-VPINNs library in terms of training time, regardless of whether \texttt{tf.float32} or \texttt{tf.float64} precision is employed.

In contrast to the linear rise in training time observed with the increase in elements for the original version of hp-VPINNs, FastVPINNs demonstrate a nearly steady training time up to 400 elements, as depicted in Figure~\ref{fig:pinn_vs_fastvpinn_time}(b) 

Section~\ref{sec:impact_of_hyperparameters} provides additional understanding regarding the influence of different hyperparameters such as the count of test functions, quadrature points, and elements on the training duration of FastVPINNs.


\subsubsection{Performance of FastVPINNs on higher frequency problems}\label{sec:spectral_fast_vpinn}
Spectral bias is a phenomenon that describes the slower learning rate of neural networks for higher frequencies compared to lower frequency solutions during the training process~\cite{cao2020understanding}. The h-refinement technique can confine test functions to a more restricted area within our domain, enabling the network to detect higher frequency solutions in that specific area~\cite{kharazmi2021hp}. In contrast to hp-VPINNs, which face longer training durations with an increased number of elements, the capability of FastVPINNs to sustain a consistent training time offers computational benefits for problems typified by higher-frequency solutions.

The aim of this section is to assess and contrast the efficacy of PINNs and FastVPINNs when dealing with solutions of varying frequencies, while keeping the number of residual points constant. In the FastVPINN scenario, the h-refinement will be adjusted according to the solution's frequency. For example, for a solution frequency of $\omega=2\pi$, a 2x2 element set-up will be used, each comprising 40x40 quadrature points. If the frequency is $\omega=4\pi$, the setup will be modified to 4x4 elements, each with 20x20 quadrature points, and so on. FastVPINNs uses five test functions in each direction. It is important to emphasize that the total count of quadrature points will remain fixed at 6,400 for all FastVPINNs configurations, equaling the number of collocation points used for PINNs training. Both models use the same parameters: 1000 Dirichlet boundary points, a neural network with three hidden layers, each containing 30 neurons, trained over 100,000 iterations with a constant learning rate of 0.001 using the Adam optimizer~\cite{Kingma2014AdamAM}

\begin{figure}[h!]
    \centering
   \includegraphics[width=0.95\linewidth]{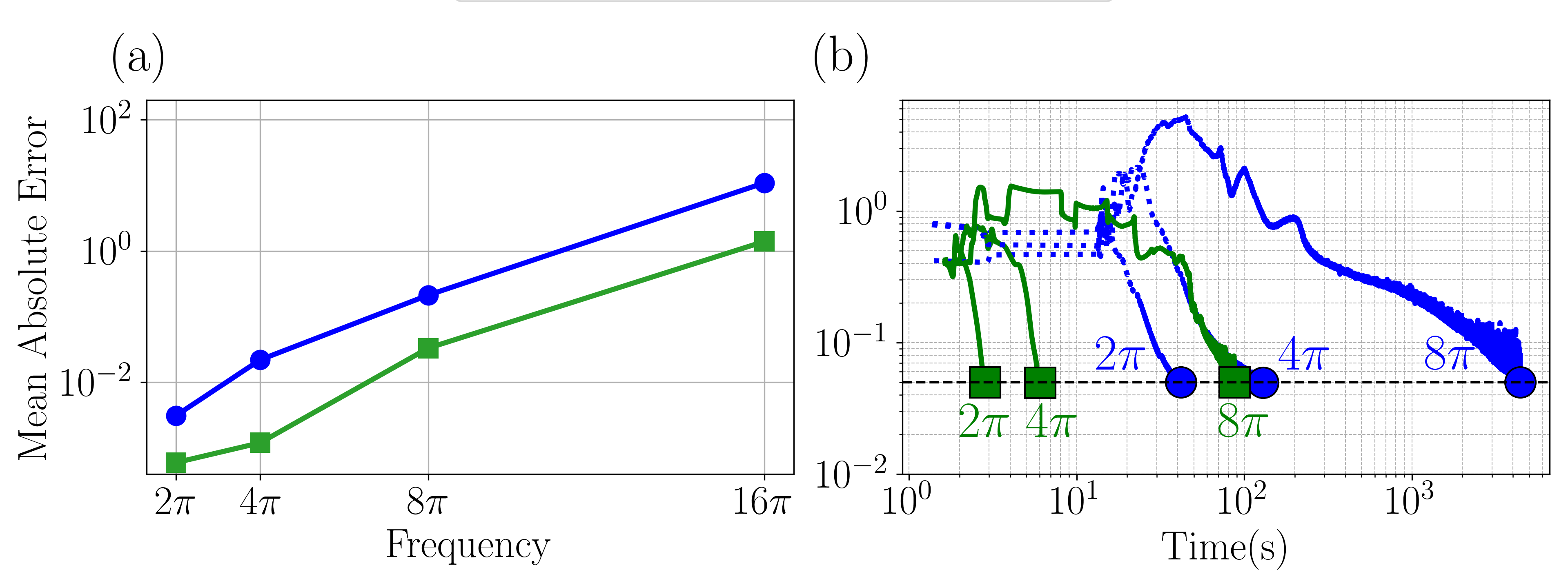}
    \caption{(a) Comparing MAE for PINNs and FastVPINNs over different frequencies. (b) time taken for PINNs and FastVPINNs to reach an MAE of $5 \times 10^{-2}$ for different frequencies.}
    \label{fig:mae_vs_time}
\end{figure}
Figure~\ref{fig:mae_vs_time}(a) demonstrates that FastVPINNs consistently outperform PINNs in Mean Absolute Error (MAE) across all tested frequencies, emphasizing the benefits of h-refinement. Furthermore, FastVPINNs reach the MAE target threshold more rapidly than traditional PINNs, as illustrated in Figure~\ref{fig:mae_vs_time}(b), highlighting their superior speed and accuracy. This also underscores the importance of carefully selecting hyperparameters such as \texttt{N\_elem}, \texttt{N\_test}, and \texttt{N\_quad} for FastVPINNs to optimize convergence and achieve efficiency and accuracy surpassing conventional PINNs.

\subsubsection{FastVPINNs on Complex Geometries}\label{sec:complex_fast_vpinn}
In order to assess the proficiency of the FastVPINN framework in handling intricate domains with a large number of elements, we used it in a 2D convection-diffusion problem ~\eqref{eq:gear_eq}on a spur gear domain~($\Omega_{\text{gear}}$) depicted in Figure~\ref{fig:gear_mesh}. The gear mesh, made up of 14,192 quadrilateral cells, was created using Gmsh(\cite{Gmsh}).

\begin{equation}
\begin{split}
    - \varepsilon \Delta u + \mathbf{b} \cdot \nabla u &= f , \quad \mathbf{x} \in \Omega_{\text{gear}}, \\
    u &= 0, \quad \mathbf{x} \in \partial\Omega_{\text{gear}},
\end{split}
\label{eq:gear_eq}
\end{equation}

where
\begin{equation*}
    f = 50\times\sin(x) + \cos(x) ; \quad \epsilon = 1,\quad \mathbf{b} = [0.1,\;\; 0]^T.  
\end{equation*}

During this experiment, we used four testing functions per direction and assigned 25 quadrature points to each component. This setup resulted in a total of 354,800 quadrature points, plus an extra 6,096 points specifically for the Dirichlet boundary ($\partial\Omega_{\text{gear}}$). The neural network was made up of 3 layers, each layer containing 50 neurons, and was launched with an initial learning rate of 0.005. This rate was adjusted through a method that reduced it by a factor of 0.99 for every 1,000 training iterations. The model was trained for a total of 150,000 iterations.
\begin{figure}[h!]
    \centering
   \includegraphics[width=0.95\linewidth]{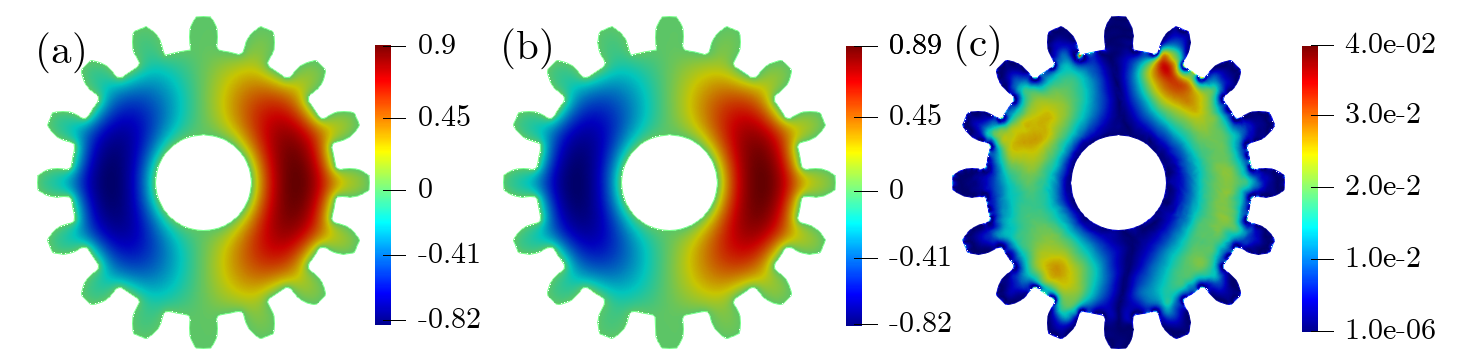}
    \caption{(a) Exact Solution obtained using FEM. (b) Predicted solution-FastVPINNs (c) Pointwise absolute error.}
    \label{fig:exact_solution_forward_complex}
\end{figure}

Figure~(\ref{fig:exact_solution_forward_complex}) shows the exact and predicted solution along with the absolute point error. It should be noted that the model training was remarkably quick, finishing in less than 35 minutes, and each iteration took approximately 13 milliseconds. This highlights the efficiency of FastVPINNs in solving problems in domains with larger element counts and skewed elements, which are challenging for the original version of hp-VPINNs. 
The comparison of prediction times between FEM and FastVPINNs for varying numbers of degrees of freedom (DOFs) is discussed in Appendix~\ref{sec:FastVPINNs vs FEM}.

\subsection{Investigating Inverse Problems with FastVPINNs}
In PINNs, inverse problems involve the use of neural networks to infer unidentified PDE parameters, aided by observed solutions at scattered data points, usually in complex physical systems. By enhancing the loss function with data from sensor points scattered across the 2D domain, the network is capable of simultaneously ascertaining the solution to the Poisson equation and the value of the parameter $\epsilon$. Consequently, the loss function for this issue is formulated as follows.
\begin{equation*}
    L_{\text{VPINN}}(W,b) = L_v + \tau L_b + \gamma L_s,
\end{equation*}
In this context, $L_s$ denotes the loss term that arises from the discrepancy between observed solutions at sensor points and their corresponding predictions from the neural network. The terms $L_v$ and $L_b$ represent the variational loss and the boundary loss, respectively, as indicated in \eqref{eqn:vpinns_final_loss_fn_abbr}. The hyperparameters $\tau$ and $\gamma$ are adjusted to suit the specific problem and the chosen architecture. This study investigates two separate categories of inverse problems related to predicting diffusion parameters: one involves uniform diffusion parameter prediction in a Poisson-2D problem, and the other involves space-dependent diffusion parameter prediction in a Convection-Diffusion-2D problem.

\begin{figure}[!htbp]
    \centering
    \includegraphics[width=0.95\linewidth]{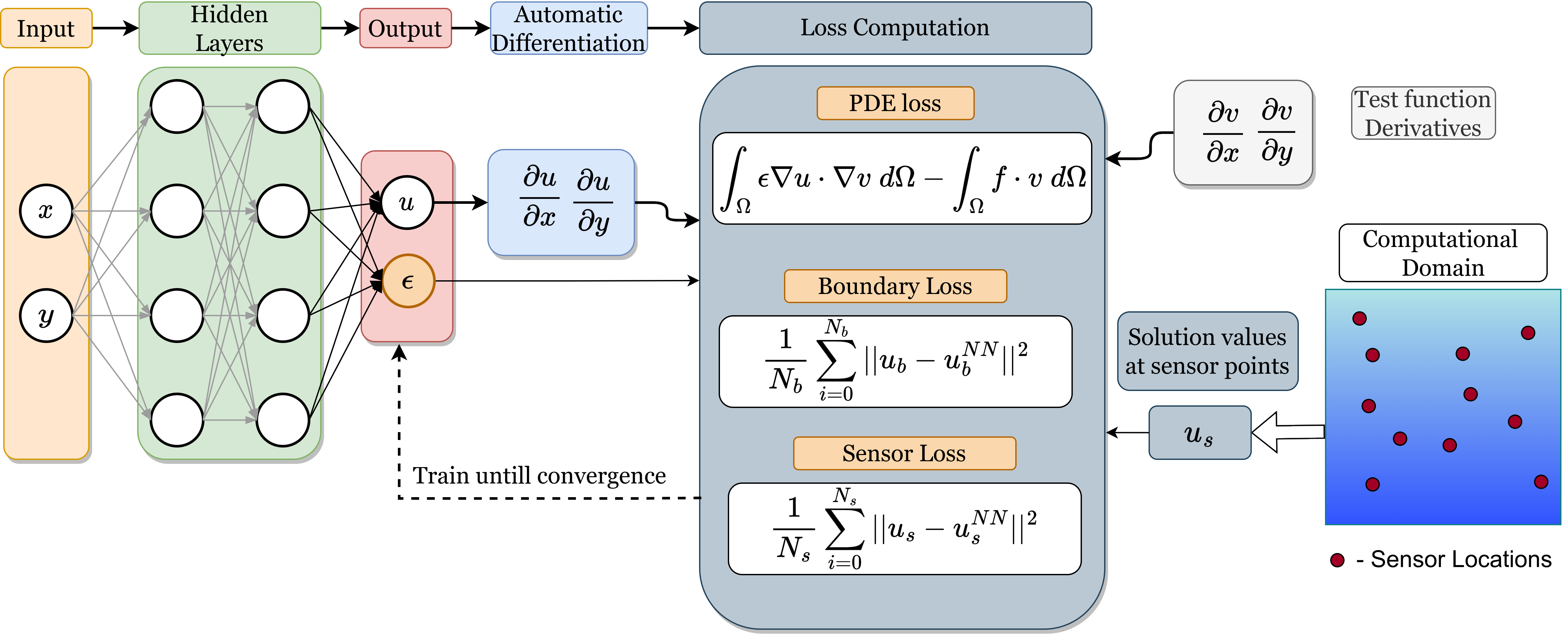}
    \caption{Schematics of VPINNs for space dependent inverse parameter modelling.}
    \label{fig:vpinn_inverse_model}
\end{figure}
\subsubsection{Estimating Constant Diffusion Parameters}\label{sec:inverse_fast_vpinns}
\begin{figure}[!htbp]
    \centering
    \includegraphics[width=0.8\linewidth]{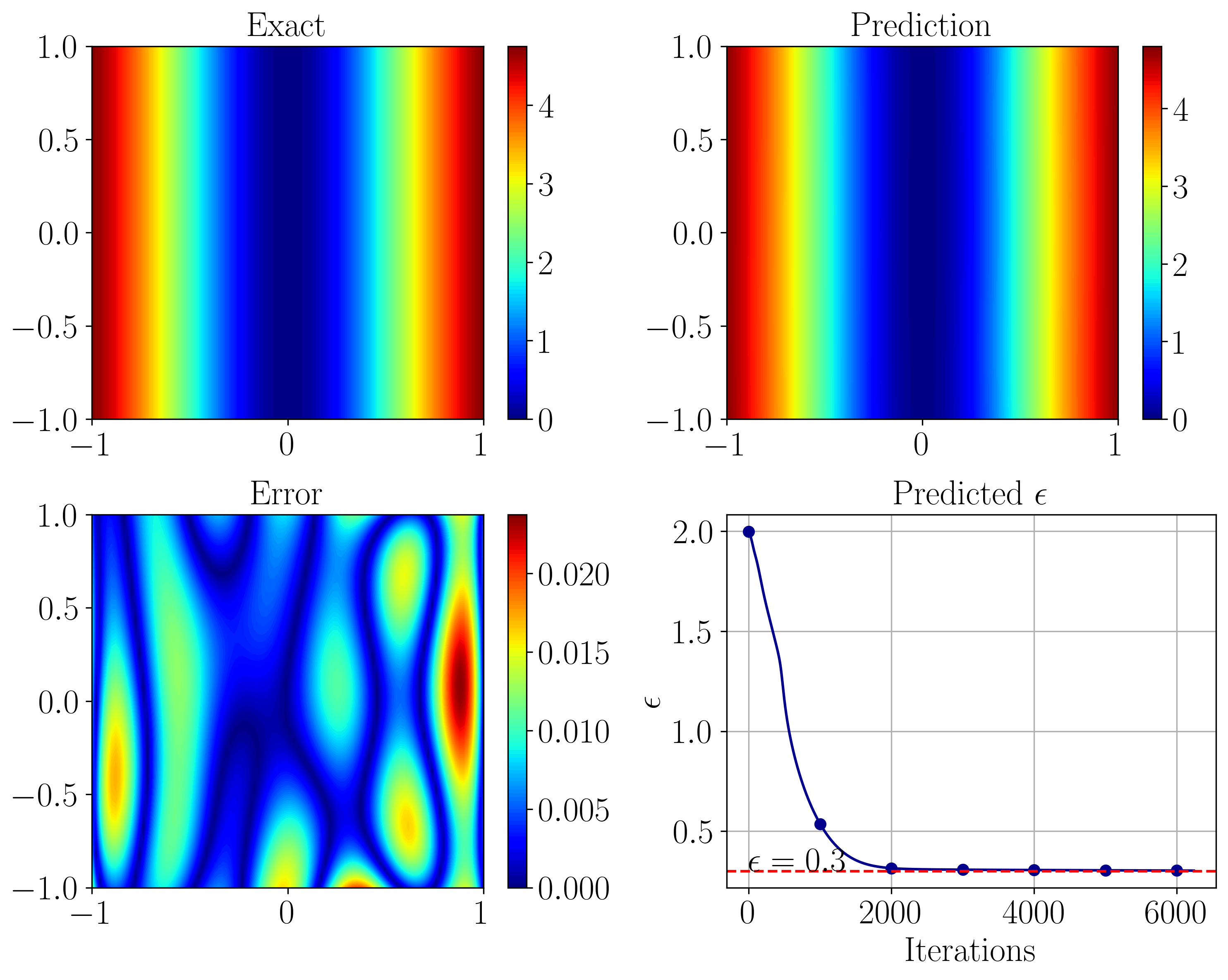}
    \caption{Inverse problem on FastVPINNs with constant diffusion parameter.}
    \label{fig:inverse_constant}
\end{figure}

In the context of uniform parameter inverse problems, we consider a two-dimensional (2D) Poisson equation as given in Eq.~(\ref{eq:poisson2D_inv_constant}). The objective is to predict the constant diffusion parameter, denoted as $\epsilon$, and the corresponding solution $u(x)$. 

\begin{align}
    \begin{split}
        -\epsilon\Delta u(x) &= f(x), \quad \ x \in \Omega = (-1, 1)^2
    \end{split}
    \label{eq:poisson2D_inv_constant}
\end{align}
Suppose the epsilon~($\epsilon$) is spatially invariant, we can incorporate it as a single trainable variable within the FastVPINNs framework. To demonstrate this approach, we built a FastVPINN with a $30 \times 3$ neural network architecture, 50 randomly distributed sensor points, a $2 \times 2$ element domain discretization, and $40 \times 40$ quadrature points per element. The exact solution to this problem is $u(x, y) = 10 \sin(x) \tanh(x) e^{-\epsilon x^2}$. The initial guess of epsilon~($\epsilon_{\text{initial}}$) is taken as 2, and the actual value of epsilon~($\epsilon_{\text{actual}}$) is 0.3. Training is carried out until it reaches the convergence criteria of $|\epsilon_{\text{predicted}} - \epsilon_{\text{actual}}| < 10^{-5}$. The network converged in 8909 epochs with a mean absolute error of the solution at $6.6 \times 10^{-2}$. The total training time was close to 18 s with a mean time per epoch of 2 ms. Figure~\ref{fig:inverse_constant} presents the exact and predicted values of both the solution and $\epsilon$.

\subsubsection{Predicting Space-Dependent Diffusion Parameters}\label{sec:domain_inverse_fast_vpinns}
\begin{figure}
    \centering
    \includegraphics[width=0.8\linewidth]{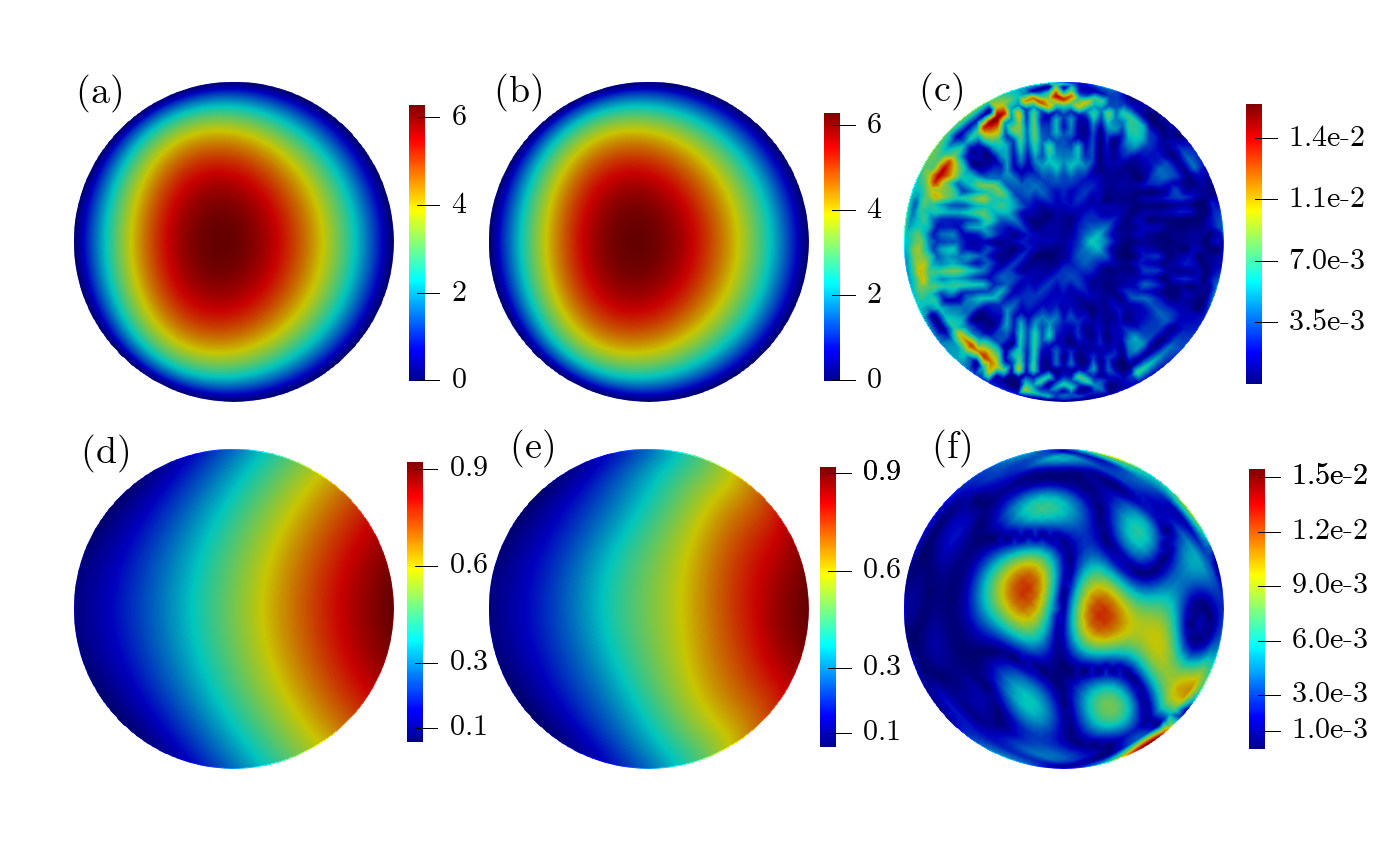}
    \caption{(a) Exact solution($u$) obtained from FEM(ParMooN). (b) Predicted Solution-FastVPINNs. (c) Absolute error: FEM vs FastVPINNs solution. (d) Exact solution diffusion parameter($\epsilon_{\text{actual}}$). (e) Predicted diffusion parameter by FastVPINNs($\epsilon_{\text{predicted}}$) (f) Absolute error: $\epsilon_{\text{actual}}$ vs $\epsilon_{\text{predicted}}$.}
    \label{fig:complex_inverse_result}
\end{figure}

This section addresses the more challenging case of predicting space-dependent parameters in partial differential equations (PDEs). As illustrated in Figure~\ref{fig:vpinn_inverse_model}, the architecture of the neural network is modified to output both the solution and the space-dependent diffusion parameter at each point, enabling their simultaneous prediction. The network is guided by sensor point data, which are substituted here with the exact solution to accurately determine both the solution and its corresponding diffusion parameter. In this example, we consider a two-dimensional (2D) convection-diffusion equation as shown in Equation~\eqref{eq:poisson2D_inv_domain}, solved on a circular domain using 1024 quadrilateral elements.

\begin{align}
    \begin{split}
        - \left( \frac{\partial}{\partial x} \left( \epsilon(x,y) \frac{\partial u}{\partial x}  \right) + \frac{\partial}{\partial y} \left( \epsilon(x,y) \frac{\partial u}{\partial y}  \right)  \right) + b_x \frac{\partial u}{\partial x} + b_y\frac{\partial u}{\partial y} &= f, \quad \ \mathbf{x} \in \Omega 
    \end{split}
    \label{eq:poisson2D_inv_domain}
\end{align}

where
\[
f = 10 ; \quad 
\epsilon_{\text{actual}}(x,y) = 0.5 \times ( \sin{x} + \cos{y} ) ; \quad
b_x = 1.0 ; \quad
b_y = 0.0 . \quad
\]
\ref{fig:complex_inverse_result} shows the reference FEM solution, FastVPINNs solution and the corresponding pointwise error for both solution and $\epsilon$. We observe that the FastVPINNs framework can reasonably predict both the solution and the spatially varying diffusion parameter with errors of the order of $\mathcal{O}(10^{-2})$ 

\subsection{Impact of Hyperparameters on FastVPINNs' Performance}\label{sec:impact_of_hyperparameters}

\begin{figure}[!htbp]
  \begin{center}
      \includegraphics[width=0.88\textwidth]{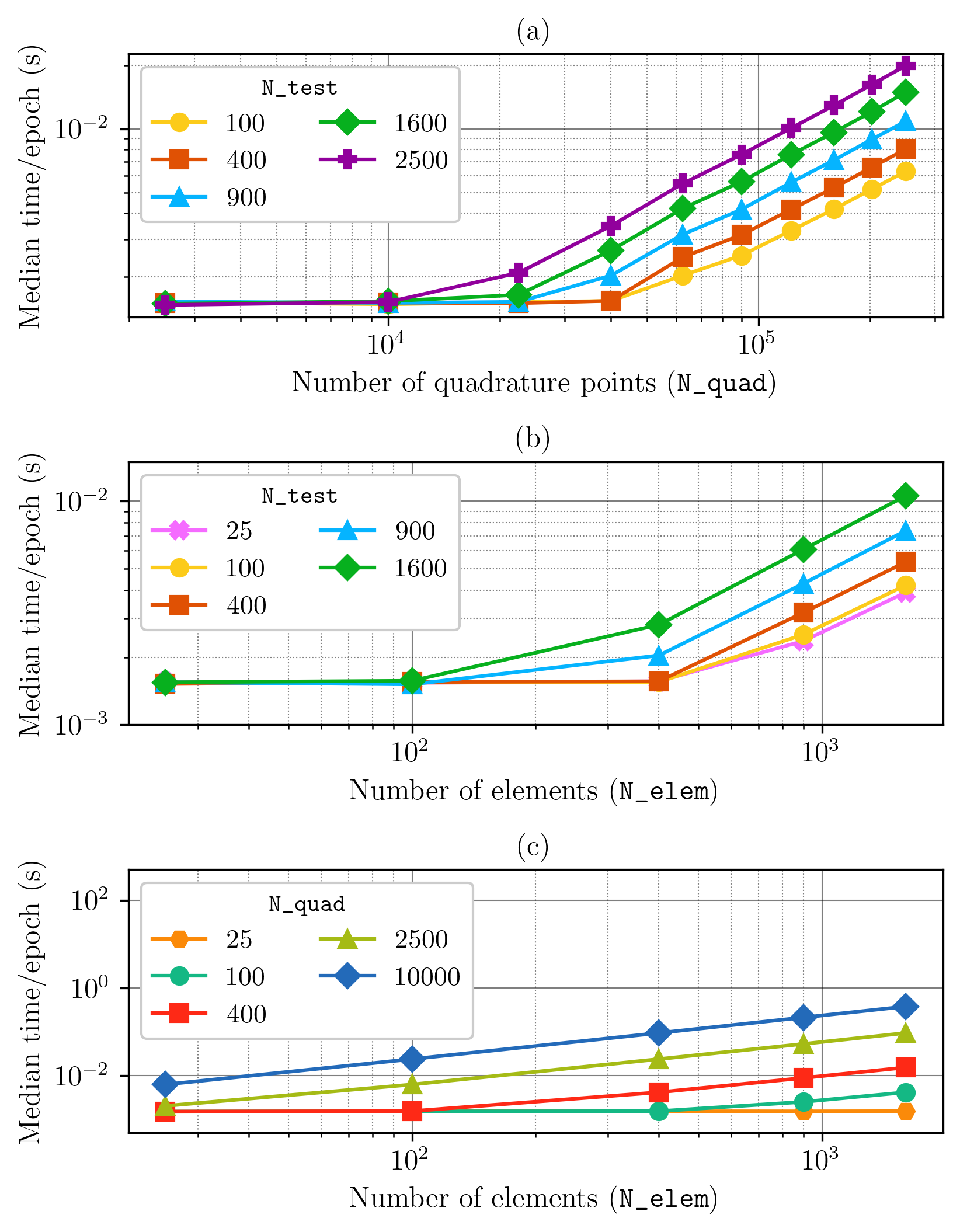} 
      \label{fig:time_vs_parameters}
  \end{center}
  
  \caption{Comparative analysis of the median training time per epoch for different hyperparameters: (a) Number of test functions~(\texttt{N\_test}) vs Number of quadrature points~(\texttt{N\_quad)} with~(\texttt{N\_elem}=1), (b) Number of Elements~(\texttt{N\_elem}) vs Number of test functions~(\texttt{N\_test}) with quadrature points fixed at 10x10~(\texttt{N\_quad}=10*10), (c) Number of elements~(\texttt{N\_elem}) vs Number of quadrature points~(\texttt{N\_quad}) with fixed number of test functions~(\texttt{N\_test}=10*10).}
  \label{fig:time-ablation}
\end{figure}

To conclude our evaluation of FastVPINNs, we investigated the impact of three critical hyperparameters on training time using a Poisson2D problem as an example. These hyperparameters are: 
\texttt{N\_quad} (quadrature points per element), 
\texttt{N\_test} (shape functions per element), and \texttt{N\_elem} (number of elements).
All experiments utilized a neural network architecture consisting of 30 layers with 3 neurons each, and conducted over 1000 training iterations. The median time per epoch was used for data visualization. Figure~\ref{fig:time-ablation}(a) illustrates the relationship between \texttt{N\_test} and \texttt{N\_quad} for a fixed \texttt{N\_elem} of 1, showing that the training time per epoch remains relatively constant regardless of the number of \texttt{N\_test} functions, up to 10,000 quadrature points.
Beyond this threshold, an increase in training time becomes noticeable, which depends on the parameters involved. Figure~\ref{fig:time-ablation}(b) shows the impact of varying \texttt{N\_test} and \texttt{N\_elem} with a fixed \texttt{N\_quad} of 100 per element. Here, the execution time remains stable across different \texttt{N\_test} functions until the number of elements reaches 100, beyond which it begins to increase. Figure~\ref{fig:time-ablation}(c) shows the effects of varying \texttt{N\_quad} and \texttt{N\_elem} 
while maintaining a constant \texttt{N\_test} of 100 per element. Figures suggest that while the overall memory occupied by the tensors might not change for certain combinations of \texttt{N\_test} and \texttt{N\_quad}, the number of quadrature points (\texttt{N\_quad}) has a more pronounced impact on training time compared to the number of test functions (\texttt{N\_test}). This observation can be attributed to the tensor-based loss computation, which performs a reduction operation along the \texttt{N\_quad} dimension, thereby underscoring the importance of \texttt{N\_quad} in influencing training time.

\section{Conclusion}
In this work, we addressed the existing challenges of hp-VPINNs, which excel in capturing high-frequency solutions but struggle with the extended training times associated with problems involving a large number of elements and complex geometries. We introduced FastVPINNs, a novel framework that employs tensor-based computations to significantly reduce training time dependence on the number of elements and to efficiently handle complex meshes. We demonstrated a 100x speedup compared to the existing hp-VPINNs implementation. Additionally, with proper hyperparameter selection, FastVPINNs can surpass state-of-the-art PINN codes in both speed and accuracy. Our method's ability to manage complex geometries with a large number of elements was showcased by solving a forward problem on a 14,000-element gear quad mesh. We successfully solved inverse problems in complex domains, as evidenced by estimating a diffusion parameter in a circular domain with 1,024 elements. This task was completed in less than 200 seconds for 100,000 epochs, which would have taken hours with the existing hp-VPINNs implementation. We also examined how hyperparameters, such as the number of test functions (\texttt{N\_test}), quadrature points (\texttt{N\_quad}), and element count (\texttt{N\_elem}), influence the training time. This analysis offers valuable insights for selecting the most suitable hyperparameters for specific problems. The versatility of FastVPINNs unlocks its potential for real-world applications in fields such as fluid dynamics, where complex geometries and massive datasets are prevalent. Looking ahead, we aim to expand the capabilities of FastVPINNs to include triangular elements and 3D domains, further broadening its impact in scientific and engineering fields.

\acks{We thank Shell Research, India, for providing partial funding for this project.  We are thankful to the MHRD Grant No. STARS-1/388 (SPADE) for partial support. We acknowledge Pratham Sunkad, for his assistance in running experiments for this paper.\\}



\bibliography{ref}

\begin{thebibliography}{31}
\providecommand{\natexlab}[1]{#1}
\providecommand{\url}[1]{\texttt{#1}}
\expandafter\ifx\csname urlstyle\endcsname\relax
  \providecommand{\doi}[1]{doi: #1}\else
  \providecommand{\doi}{doi: \begingroup \urlstyle{rm}\Url}\fi

\bibitem[Abadi et~al.(2015)Abadi, Agarwal, Barham, Brevdo, Chen, Citro, Corrado, Davis, Dean, Devin, Ghemawat, Goodfellow, Harp, Irving, Isard, Jia, Jozefowicz, Kaiser, Kudlur, Levenberg, Man\'{e}, Monga, Moore, Murray, Olah, Schuster, Shlens, Steiner, Sutskever, Talwar, Tucker, Vanhoucke, Vasudevan, Vi\'{e}gas, Vinyals, Warden, Wattenberg, Wicke, Yu, and Zheng]{tensorflow2015-whitepaper}
M.~Abadi, A.~Agarwal, P.~Barham, E.~Brevdo, Z.~Chen, C.~Citro, G.~S. Corrado, A.~Davis, J.~Dean, M.~Devin, S.~Ghemawat, I.~Goodfellow, A.~Harp, G.~Irving, M.~Isard, Y.~Jia, R.~Jozefowicz, L.~Kaiser, M.~Kudlur, J.~Levenberg, D.~Man\'{e}, R.~Monga, S.~Moore, D.~Murray, C.~Olah, M.~Schuster, J.~Shlens, B.~Steiner, I.~Sutskever, K.~Talwar, P.~Tucker, V.~Vanhoucke, V.~Vasudevan, F.~Vi\'{e}gas, O.~Vinyals, P.~Warden, M.~Wattenberg, M.~Wicke, Y.~Yu, and X.~Zheng.
\newblock {TensorFlow}: Large-scale machine learning on heterogeneous systems, 2015.
\newblock URL \url{https://www.tensorflow.org/}.
\newblock Software available from tensorflow.org.

\bibitem[Abueidda et~al.(2021)Abueidda, Lu, and Koric]{abueidda2021meshless}
D.~W. Abueidda, Q.~Lu, and S.~Koric.
\newblock {Meshless physics-informed deep learning method for three-dimensional solid mechanics}.
\newblock \emph{International Journal for Numerical Methods in Engineering}, 122\penalty0 (23):\penalty0 7182--7201, 2021.

\bibitem[Baker et~al.(2019)Baker, Alexander, Bremer, Hagberg, Kevrekidis, Najm, Parashar, Patra, Sethian, Wild, Willcox, and Lee]{osti_1478744}
N.~Baker, F.~Alexander, T.~Bremer, A.~Hagberg, Y.~Kevrekidis, H.~Najm, M.~Parashar, A.~Patra, J.~Sethian, S.~Wild, K.~Willcox, and S.~Lee.
\newblock Workshop report on basic research needs for scientific machine learning: Core technologies for artificial intelligence.
\newblock 2 2019.
\newblock \doi{10.2172/1478744}.
\newblock URL \url{https://www.osti.gov/biblio/1478744}.

\bibitem[Bangerth et~al.(2007)Bangerth, Hartmann, and Kanschat]{bangerth2007deal}
W.~Bangerth, R.~Hartmann, and G.~Kanschat.
\newblock {deal. II—a general-purpose object-oriented finite element library}.
\newblock \emph{ACM Transactions on Mathematical Software (TOMS)}, 33\penalty0 (4):\penalty0 24--es, 2007.

\bibitem[Cai et~al.(2021)Cai, Mao, Wang, Yin, and Karniadakis]{cai2021physics}
S.~Cai, Z.~Mao, Z.~Wang, M.~Yin, and G.~E. Karniadakis.
\newblock {Physics-informed neural networks (PINNs) for fluid mechanics: A review}.
\newblock \emph{Acta Mechanica Sinica}, 37\penalty0 (12):\penalty0 1727--1738, 2021.

\bibitem[Cao et~al.(2020)Cao, Fang, Wu, Zhou, and Gu]{cao2020understanding}
Y.~Cao, Z.~Fang, Y.~Wu, D.-X. Zhou, and Q.~Gu.
\newblock Towards understanding the spectral bias of deep learning, 2020.

\bibitem[Cuomo et~al.(2022)Cuomo, Di~Cola, Giampaolo, Rozza, Raissi, and Piccialli]{cuomo2022scientific}
S.~Cuomo, V.~S. Di~Cola, F.~Giampaolo, G.~Rozza, M.~Raissi, and F.~Piccialli.
\newblock {Scientific machine learning through physics--informed neural networks: Where we are and what’s next}.
\newblock \emph{Journal of Scientific Computing}, 92\penalty0 (3):\penalty0 88, 2022.

\bibitem[Eivazi et~al.(2022)Eivazi, Tahani, Schlatter, and Vinuesa]{eivazi2022physics}
H.~Eivazi, M.~Tahani, P.~Schlatter, and R.~Vinuesa.
\newblock Physics-informed neural networks for solving {R}eynolds-averaged {N}avier--{S}tokes equations.
\newblock \emph{Physics of Fluids}, 34\penalty0 (7), 2022.

\bibitem[Ganesan and Shah(2020)]{ganesan2020sparsh}
S.~Ganesan and M.~Shah.
\newblock {SParSH-AMG: A library for hybrid CPU-GPU algebraic multigrid and preconditioned iterative methods}.
\newblock \emph{arXiv preprint arXiv:2007.00056}, 2020.

\bibitem[Ganesan and Tobiska(2017)]{ganesan2017finite}
S.~Ganesan and L.~Tobiska.
\newblock \emph{Finite elements: Theory and algorithms}.
\newblock Cambridge University Press, 2017.

\bibitem[Geuzaine and Remacle(2009)]{Gmsh}
C.~Geuzaine and J.-F. Remacle.
\newblock {Gmsh: A 3--{D} finite element mesh generator with built-in pre- and post-processing facilities}.
\newblock \emph{International Journal for Numerical Methods in Engineering}, 79\penalty0 (11):\penalty0 1309--1331, 2009.
\newblock \doi{https://doi.org/10.1002/nme.2579}.
\newblock URL \url{https://onlinelibrary.wiley.com/doi/abs/10.1002/nme.2579}.

\bibitem[Haghighat et~al.(2021)Haghighat, Raissi, Moure, Gomez, and Juanes]{haghighat2021physics}
E.~Haghighat, M.~Raissi, A.~Moure, H.~Gomez, and R.~Juanes.
\newblock A physics-informed deep learning framework for inversion and surrogate modeling in solid mechanics.
\newblock \emph{Computer Methods in Applied Mechanics and Engineering}, 379:\penalty0 113741, 2021.

\bibitem[Kharazmi(2023)]{hp_vpinns_github}
E.~Kharazmi.
\newblock {hp-VPINNs}: High-performance variational physics-informed neural networks, 2023.
\newblock URL \url{https://github.com/ehsankharazmi/hp-VPINNs}.
\newblock Accessed: 2023-12-01.

\bibitem[Kharazmi et~al.(2019)Kharazmi, Zhang, and Karniadakis]{kharazmi2019variational}
E.~Kharazmi, Z.~Zhang, and G.~E. Karniadakis.
\newblock {Variational physics-informed neural networks for solving partial differential equations}.
\newblock \emph{arXiv preprint arXiv:1912.00873}, 2019.

\bibitem[Kharazmi et~al.(2021)Kharazmi, Zhang, and Karniadakis]{kharazmi2021hp}
E.~Kharazmi, Z.~Zhang, and G.~E. Karniadakis.
\newblock {hp-VPINNs: Variational physics-informed neural networks with domain decomposition}.
\newblock \emph{Computer Methods in Applied Mechanics and Engineering}, 374:\penalty0 113547, 2021.

\bibitem[Khodayi-Mehr and Zavlanos(2020)]{khodayi2020varnet}
R.~Khodayi-Mehr and M.~Zavlanos.
\newblock Varnet: Variational neural networks for the solution of partial differential equations.
\newblock In \emph{Learning for Dynamics and Control}, pages 298--307. PMLR, 2020.

\bibitem[Kingma and Ba(2014)]{Kingma2014AdamAM}
D.~P. Kingma and J.~Ba.
\newblock Adam: A method for stochastic optimization.
\newblock \emph{CoRR}, abs/1412.6980, 2014.
\newblock URL \url{https://api.semanticscholar.org/CorpusID:6628106}.

\bibitem[Lagaris et~al.(1998)Lagaris, Likas, and Fotiadis]{712178}
I.~Lagaris, A.~Likas, and D.~Fotiadis.
\newblock Artificial neural networks for solving ordinary and partial differential equations.
\newblock \emph{IEEE Transactions on Neural Networks}, 9\penalty0 (5):\penalty0 987--1000, 1998.
\newblock \doi{10.1109/72.712178}.

\bibitem[Liu and Wu(2023)]{LIU2023102051}
C.~Liu and H.~Wu.
\newblock {cv-PINN: Efficient learning of variational physics-informed neural network with domain decomposition}.
\newblock \emph{Extreme Mechanics Letters}, 63:\penalty0 102051, 2023.
\newblock ISSN 2352-4316.
\newblock \doi{https://doi.org/10.1016/j.eml.2023.102051}.
\newblock URL \url{https://www.sciencedirect.com/science/article/pii/S2352431623000974}.

\bibitem[Lu et~al.(2021{\natexlab{a}})Lu, Meng, Mao, and Karniadakis]{lu2021deepxde}
L.~Lu, X.~Meng, Z.~Mao, and G.~E. Karniadakis.
\newblock {DeepXDE: A deep learning library for solving differential equations}.
\newblock \emph{SIAM review}, 63\penalty0 (1):\penalty0 208--228, 2021{\natexlab{a}}.

\bibitem[Lu et~al.(2021{\natexlab{b}})Lu, Pestourie, Yao, Wang, Verdugo, and Johnson]{lu2021physics}
L.~Lu, R.~Pestourie, W.~Yao, Z.~Wang, F.~Verdugo, and S.~G. Johnson.
\newblock Physics-informed neural networks with hard constraints for inverse design.
\newblock \emph{SIAM Journal on Scientific Computing}, 43\penalty0 (6):\penalty0 B1105--B1132, 2021{\natexlab{b}}.

\bibitem[Mao et~al.(2020)Mao, Jagtap, and Karniadakis]{mao2020physics}
Z.~Mao, A.~D. Jagtap, and G.~E. Karniadakis.
\newblock Physics-informed neural networks for high-speed flows.
\newblock \emph{Computer Methods in Applied Mechanics and Engineering}, 360:\penalty0 112789, 2020.

\bibitem[{NVIDIA}(2023)]{nvidia_modulus_docs}
{NVIDIA}.
\newblock {NVIDIA Modulus Documentation}.
\newblock \url{https://docs.nvidia.com/deeplearning/modulus/modulus-sym/index.html}, 2023.
\newblock Accessed: 2023-03-10.

\bibitem[NVIDIA Modulus()]{modulus}
NVIDIA Modulus.
\newblock \url{https://developer.nvidia.com/modulus}. Last accessed Jan 01, 2024.

\bibitem[Psaros et~al.(2023)Psaros, Meng, Zou, Guo, and Karniadakis]{psaros2023uncertainty}
A.~F. Psaros, X.~Meng, Z.~Zou, L.~Guo, and G.~E. Karniadakis.
\newblock Uncertainty quantification in scientific machine learning: Methods, metrics, and comparisons.
\newblock \emph{Journal of Computational Physics}, 477:\penalty0 111902, 2023.

\bibitem[Radin et~al.(2023)Radin, Klinkel, and Altay]{radin2023effects}
N.~Radin, S.~Klinkel, and O.~Altay.
\newblock Effects of variational formulations on physics-informed neural network performance in solid mechanics.
\newblock \emph{PAMM}, page e202300222, 2023.

\bibitem[Raissi et~al.(2019)Raissi, Perdikaris, and Karniadakis]{raissi2019physics}
M.~Raissi, P.~Perdikaris, and G.~E. Karniadakis.
\newblock {Physics-informed neural networks: A deep learning framework for solving forward and inverse problems involving nonlinear partial differential equations}.
\newblock \emph{Journal of Computational physics}, 378:\penalty0 686--707, 2019.

\bibitem[Smith et~al.(2021)Smith, Ross, Azizzadenesheli, and Muir]{10.1093/gji/ggab309}
J.~D. Smith, Z.~E. Ross, K.~Azizzadenesheli, and J.~B. Muir.
\newblock {{HypoSVI: Hypocentre inversion with Stein variational inference and physics informed neural networks}}.
\newblock \emph{Geophysical Journal International}, 228\penalty0 (1):\penalty0 698--710, 08 2021.
\newblock ISSN 0956-540X.
\newblock \doi{10.1093/gji/ggab309}.
\newblock URL \url{https://doi.org/10.1093/gji/ggab309}.

\bibitem[Wilbrandt et~al.(2017)Wilbrandt, Bartsch, Ahmed, Alia, Anker, Blank, Caiazzo, Ganesan, Giere, Matthies, et~al.]{wilbrandt2017parmoon}
U.~Wilbrandt, C.~Bartsch, N.~Ahmed, N.~Alia, F.~Anker, L.~Blank, A.~Caiazzo, S.~Ganesan, S.~Giere, G.~Matthies, et~al.
\newblock Par{M}oo{N}—a modernized program package based on mapped finite elements.
\newblock \emph{Computers \& Mathematics with Applications}, 74\penalty0 (1):\penalty0 74--88, 2017.

\bibitem[Yang and Foster(2021)]{yang2021hp}
M.~Yang and J.~T. Foster.
\newblock {hp-Variational Physics-Informed Neural Networks for Nonlinear Two-Phase Transport in Porous Media}.
\newblock \emph{Journal of Machine Learning for Modeling and Computing}, 2\penalty0 (2), 2021.

\bibitem[Zhang et~al.(2022)Zhang, Dao, Karniadakis, and Suresh]{zhang2022analyses}
E.~Zhang, M.~Dao, G.~E. Karniadakis, and S.~Suresh.
\newblock Analyses of internal structures and defects in materials using physics-informed neural networks.
\newblock \emph{Science advances}, 8\penalty0 (7):\penalty0 eabk0644, 2022.

\end{thebibliography}

\newpage

\appendix
\section{FastVPINNs Methodology}
\subsection{Bilinear Transformation}\label{appendix: bilinear}
Let \( b_0(-1,-1) \), \( b_1(1,-1) \), \( b_2(1,1) \), \( b_3(-1,1) \) be the vertices of the reference element \(\hat{K}\), see Figure~\ref{fig:bilinear Transformation}.
For any function \( u(X) \), we denote \( u(X) = u(F_k(\hat{X})) = \hat{u}(\hat{X}) \). Further, the derivatives of the function, \( \partial u/\partial x \) and \( \partial u/\partial y \) on the original element can be obtained in terms of the derivatives defined on the refernce element~\( \partial \hat{u}/\partial \xi \) and \( \partial \hat{u}/\partial \eta \) as follows:
\begin{align}
\hat{u}(\hat{X}) &= \hat{u}(F_k^{-1}(X)) = u(X), \\
\frac{\partial \hat{u}}{\partial \xi} &= \frac{\partial u}{\partial x} \frac{\partial x}{\partial \xi} + \frac{\partial u}{\partial y} \frac{\partial y}{\partial \xi}, \\
\frac{\partial \hat{u}}{\partial \eta} &= \frac{\partial u}{\partial x} \frac{\partial x}{\partial \eta} + \frac{\partial u}{\partial y} \frac{\partial y}{\partial \eta}.
\end{align}
we can express this relation as,
\begin{align*}
\renewcommand{\arraystretch}{1.5} 
\begin{bmatrix}
\frac{\partial \hat{u}}{\partial \xi} \\
\frac{\partial \hat{u}}{\partial \eta}
\end{bmatrix} &=
\renewcommand{\arraystretch}{1.5} 
\begin{bmatrix}
(x_{c1} + x_{c3}\eta) & (y_{c1} + y_{c3}\eta) \\
(x_{c2} + x_{c3}\xi) & (y_{c2} + y_{c3}\xi)
\end{bmatrix}
\renewcommand{\arraystretch}{1.5} 
\begin{bmatrix}
\frac{\partial u}{\partial x} \\
\frac{\partial u}{\partial y}
\end{bmatrix}.
\end{align*}
where, 
\begin{align*}
x_{c0} &= \frac{(x_0 + x_1 + x_2 + x_3)}{4}, & x_{c1} &= \frac{(-x_0 + x_1 + x_2 - x_3)}{4}, \\
x_{c2} &= \frac{(-x_0 - x_1 + x_2 + x_3)}{4}, & x_{c3} &= \frac{(x_0 - x_1 + x_2 - x_3)}{4}, \\
y_{c0} &= \frac{(y_0 + y_1 + y_2 + y_3)}{4}, & y_{c1} &= \frac{(-y_0 + y_1 + y_2 - y_3)}{4}, \\
y_{c2} &= \frac{(-y_0 - y_1 + y_2 + y_3)}{4}, & y_{c3} &= \frac{(y_0 - y_1 + y_2 - y_3)}{4},
\end{align*}

Finally, we have
Finally, we have
\begin{align*}
\renewcommand{\arraystretch}{1.5} 
\begin{bmatrix}
\frac{\partial u}{\partial x} \\
\frac{\partial u}{\partial y}
\end{bmatrix} &=
\renewcommand{\arraystretch}{1.5} 
\frac{1}{D}
\begin{bmatrix}
(y_{c2} + y_{c3}\xi) & -(y_{c1} + y_{c3}\eta) \\
-(x_{c2} + x_{c3}\eta) & (x_{c1} + x_{c3}\xi)
\end{bmatrix}
\renewcommand{\arraystretch}{1.5} 
\begin{bmatrix}
\frac{\partial \hat{u}}{\partial \xi} \\
\frac{\partial \hat{u}}{\partial \eta}
\end{bmatrix},
\end{align*}

where, \( D \) is the determinant of the Jacobian matrix 
\newpage

\subsection{Algorithm : Premultiplier Assembly for Algorithm 2}

\noindent\rule{\textwidth}{1pt}  
Algorithm : hp-VPINNs vectorised code - Premultiplier Assembly  \newline
\noindent\rule{\textwidth}{1pt}  
\begin{lstlisting}[language=Python]
# Compute the prematrix multipliers for shapefunctions and gradients
# Shape: (N_test, N_quad)
for j in range(N_test):
  for q in range(N_quad):   `
    V_k[j][q]  = quad_wt[q] * v_k[j][q]  # Premult. shape mat 
    V_x[j][q]  = quad_wt[q] * v_kx[j][q]  # Premult. grad x mat
    V_y[j][q]  = quad_wt[q] * v_ky[j][q]  # Premult. grad y mat

# Compute Forcing Matrix
# Shape (N_Elem, N_Test)
for n in range(N_elem):
  for j in range(N_test):
    for q in range(N_quad):
      # xi and eta are quadrature co-ordinates in reference domain
      x,y = fe.get_actual_coordinates(n,xi[q],eta[q])
      F[n][j] += j[n][q] * qw[q] * force_fn(x,y) * test[j][q]

# Collect the jacobian terms and stack them
# Shape: (N_quad, N_elem) 
for q in range(N_quad):
  for n in range(N_elem):
    J_x[q][n] = j[n][q] / j_x[n][q] # Jacobian x matrix
    J_y[q][n] = j[n][q] / j_y[n][q] # Jacobian y matrix

\end{lstlisting}
\noindent\rule{\textwidth}{1pt}  

\newpage
\newpage
\section{Results}
\subsection{Effect of h- and p-refinement in FastVPINNs}\label{sec:appendix_hp_refinement}
\begin{figure}[h!]
  \begin{center}
      \includegraphics{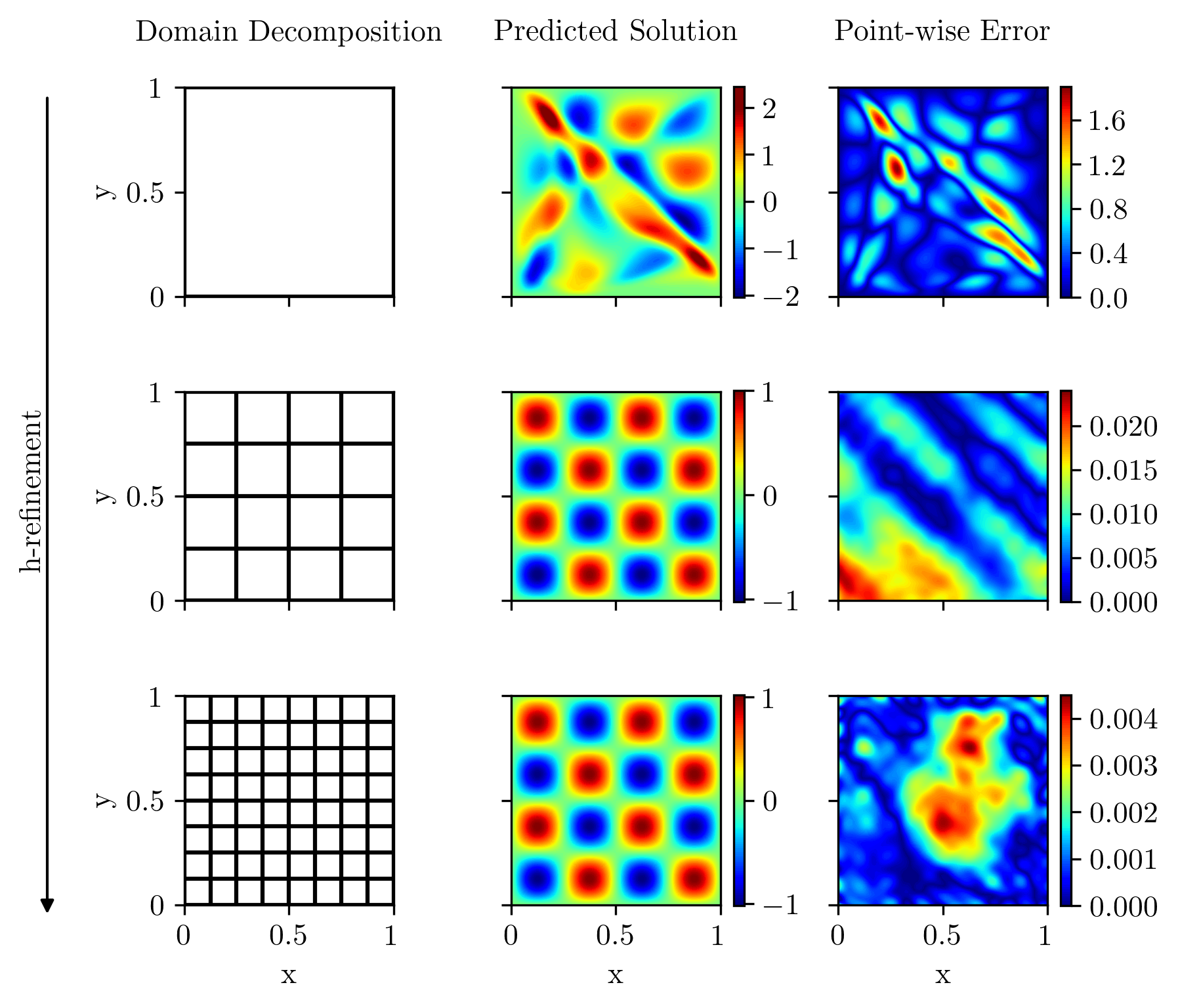} 
  \end{center}
  
  \caption{Effect of h-refinement on the accuracy of solving the two-dimensional Poisson equation. The first, second, and third columns are the domain decomposition, FastVPINNs solution, and pointwise test error, respectively. From top to bottom: \texttt{N\_elem} = 1, \texttt{N\_elem} = 16 and \texttt{N\_elem} = 64. For each element, we use $80\times 80$ quadrature points and $5$ test functions in each direction.}
  \label{fig:h-refinement}
\end{figure}
\begin{figure}[h!]
  \centering
  \includegraphics{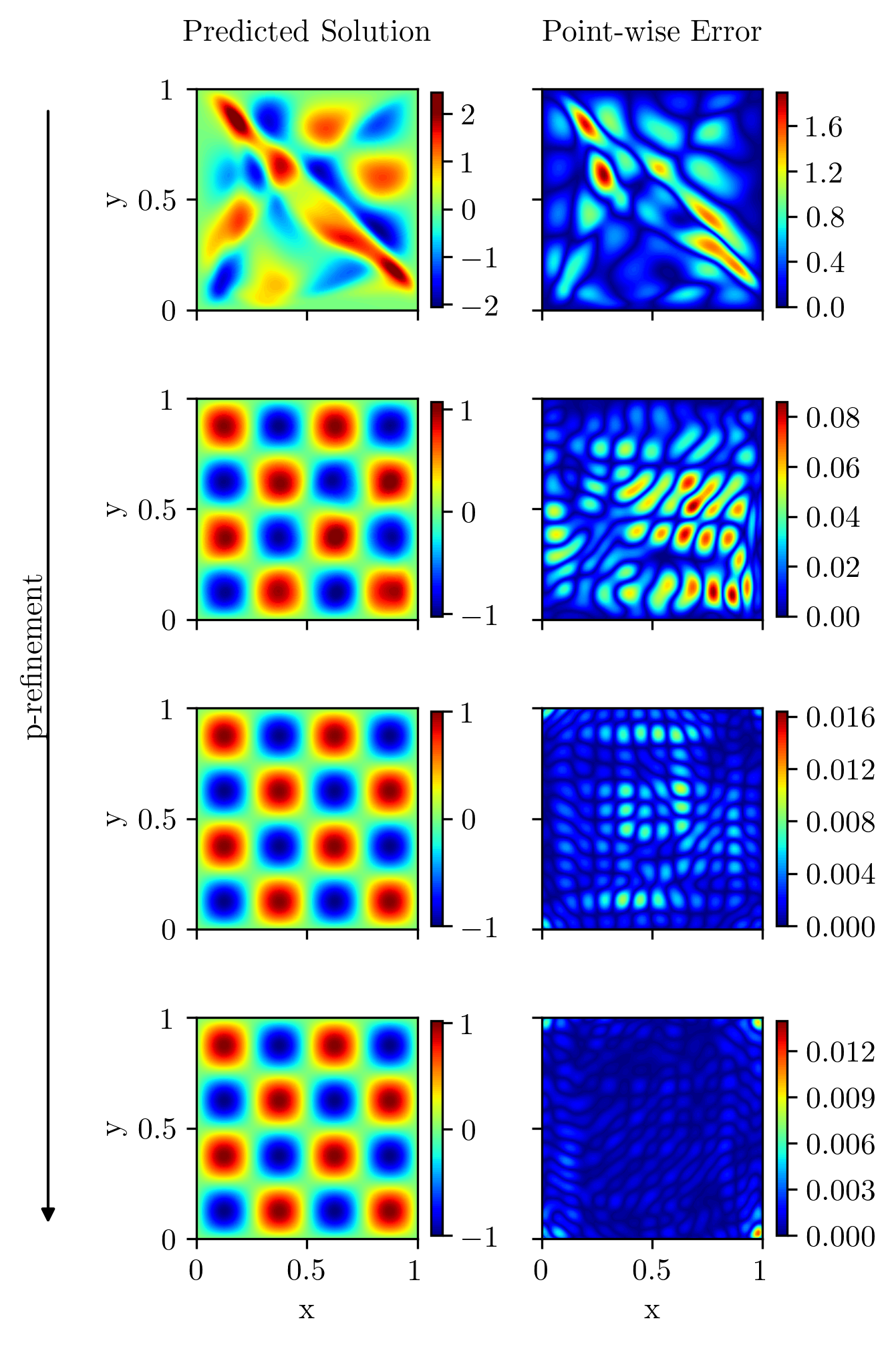} 
  \caption{Effect of p-refinement on the accuracy of solution of two-dimensional Poisson's equation on 1 element. The first and second columns are the FastVPINNs solution and point-wise test error, respectively. From top to bottom: $\texttt{N\_test} = 5\times 5$, $\texttt{N\_test} = 10\times 10$, $\texttt{N\_test} = 15\times 15$ and $\texttt{N\_test} = 20\times 20$. The element has $80 \times 80$ total quadrature points.}
  \label{fig:p-refinement}
\end{figure}

\newpage

\subsection{Time Comparison - FastVPINNs vs FEM}\label{sec:FastVPINNs vs FEM}
\begin{figure}[h!]
  \centering
  \includegraphics[width=0.6\textwidth]{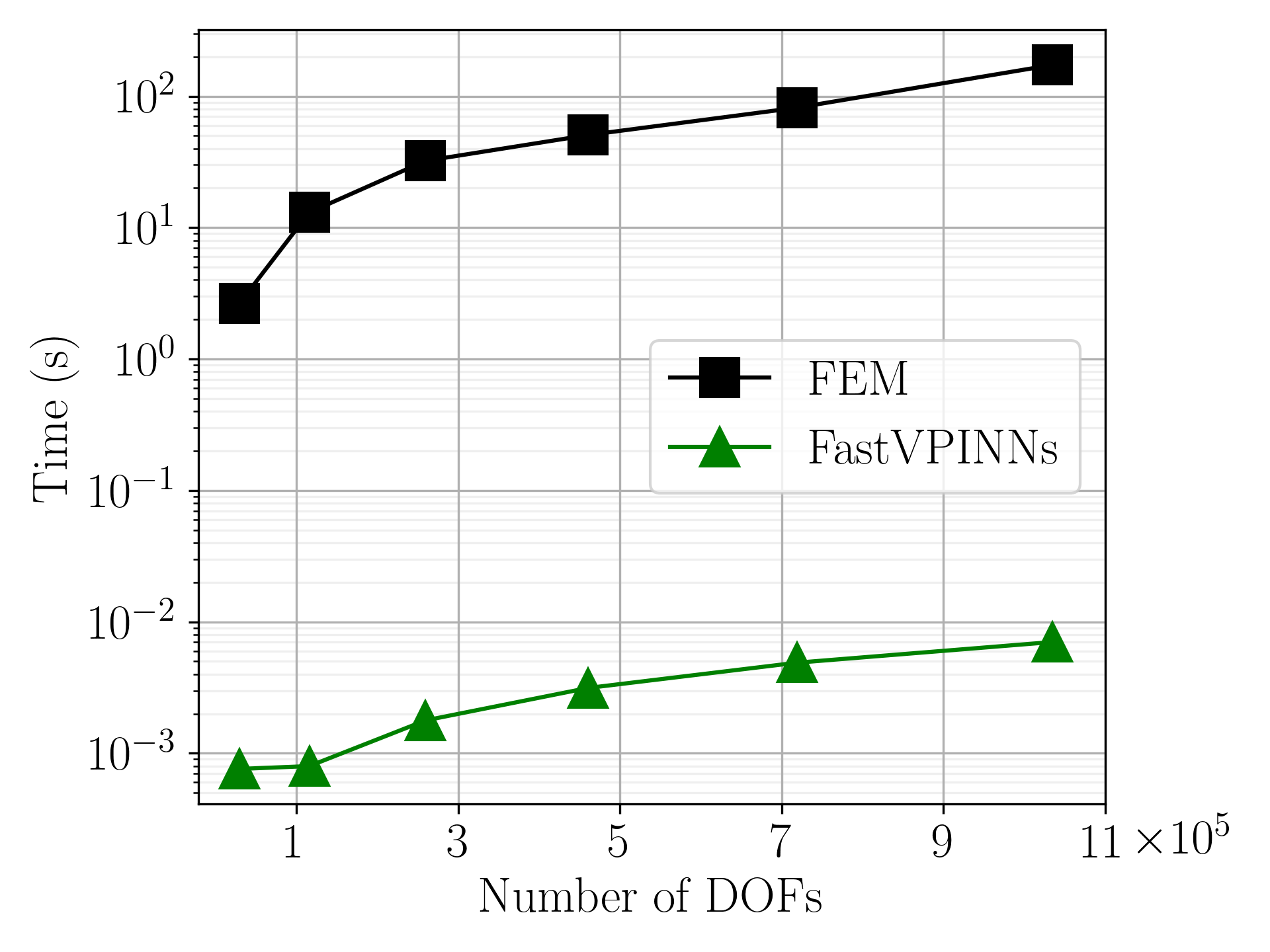}
  \caption{Comparision of time taken for prediction by FEM and PINNs for varying number of quadrature points.}
  \label{fig:FastVPINNs vs PINNs}
\end{figure}

The Table~\ref{tab:time_prediction} and Figure~\ref{fig:FastVPINNs vs PINNs} below represents the time required for the prediction of the solution using PINNs and FEM. It's important to note that for FastVPINNs, this time does not include the training time.
\begin{table}[!t]
\caption{Time Taken (in s) for Prediction using FEM and PINNs}
\centering
\begin{tabular}{|c|c|c|}
\hline
\textbf{N\_DOF} & \textbf{FEM} & \textbf{PINNs} \\
\hline
29302   & 2.638 & 7.807e-4 \\
115868  & 13.0352 & 7.978e-4 \\
259698  & 32.220 & 1.777e-3  \\
460792  & 50.680  & 3.139e-3  \\
719150  & 81.784 & 4.890e-3  \\
1034772 & 172.967 & 7.023e-3  \\ 
\hline
\end{tabular}
\label{tab:time_prediction}
\end{table}

\end{document}